\documentclass[conference]{IEEEtran}
\IEEEoverridecommandlockouts

\usepackage{cite}
\usepackage{amsmath,amssymb,amsfonts}
\usepackage{algorithmic}
\usepackage{graphicx}
\usepackage{textcomp}
\usepackage{xcolor}

\usepackage{flushend}
\usepackage{caption}
\usepackage{tabularx}
\usepackage{makecell}
\usepackage{subcaption}
\usepackage{amsthm}
\usepackage{algorithm}
\usepackage{multirow}
\usepackage{color}
\usepackage{booktabs}
\usepackage{float}
\usepackage{balance}
\usepackage{xspace}
\usepackage{threeparttable}
\usepackage{url}
\usepackage{enumitem}
\usepackage{amssymb}
\usepackage{marvosym}
\usepackage[
    colorlinks=true,
    linkcolor=red,      
    urlcolor=blue,      
    citecolor=green,    
    linkbordercolor=red 
]{hyperref}

\newtheorem{Def}{Definition}

\newtheorem*{Pro*}{Problem}
\newcommand{\model}{UMGAD\xspace}

\def\BibTeX{{\rm B\kern-.05em{\sc i\kern-.025em b}\kern-.08em
    T\kern-.1667em\lower.7ex\hbox{E}\kern-.125emX}}
\begin{document}

\title{UMGAD: Unsupervised Multiplex Graph Anomaly Detection}

\author{
\IEEEauthorblockN{
Xiang Li\textsuperscript{1}, Jianpeng Qi\textsuperscript{1}, Zhongying Zhao\textsuperscript{2}, Guanjie Zheng\textsuperscript{3}, Lei Cao\textsuperscript{4}, Junyu Dong\textsuperscript{1}, Yanwei Yu\textsuperscript{1}\textsuperscript{\Letter}}\\
\IEEEauthorblockA{
\textsuperscript{1}Faculty of Information Science and Engineering, Ocean University of China, Qingdao, China\\
\textsuperscript{2}College of Computer Science and Engineering, Shandong University of Science and Technology, Qingdao, China\\
\textsuperscript{3}Department of Computer Science and Engineering, Shanghai Jiao Tong University, Shanghai, China\\
\textsuperscript{4}Department of Computer Science, The University of Arizona, Tucson, USA\\
lixiang1202@stu.ouc.edu.cn, zyzhao@sdust.edu.cn, gjzheng@sjtu.edu.cn, lcao@csail.mit.edu, \\\{qijianpeng, dongjunyu, yuyanwei\}@ouc.edu.cn}\\
\thanks{Corresponding author: Yanwei Yu.}
}

\maketitle

\begin{abstract}
Graph anomaly detection (GAD) is a critical task in graph machine learning, with the primary objective of identifying anomalous nodes that deviate significantly from the majority. This task is widely applied in various real-world scenarios, including fraud detection and social network analysis.
 However, existing GAD methods still face two major challenges: (1) They are often limited to detecting anomalies in single-type interaction graphs and struggle with multiple interaction types in multiplex heterogeneous graphs. (2) In unsupervised scenarios, selecting appropriate anomaly score thresholds remains a significant challenge for accurate anomaly detection. To address the above challenges, we propose a novel \underline{\textbf{U}}nsupervised \underline{\textbf{M}}ultiplex \underline{\textbf{G}}raph \underline{\textbf{A}}nomaly \underline{\textbf{D}}etection method, named \model.
 We first learn multi-relational correlations among nodes in multiplex heterogeneous graphs and capture anomaly information during node attribute and structure reconstruction through graph-masked autoencoder (GMAE). Then, to further extract abnormal information, we generate attribute-level and subgraph-level augmented-view graphs, respectively, and perform attribute and structure reconstruction through GMAE. Finally, we learn to optimize node attributes and structural features through contrastive learning between original-view and augmented-view graphs to improve the model's ability to capture anomalies. Meanwhile, we propose a new anomaly score threshold selection strategy, which allows the model to be independent of ground truth information in real unsupervised scenarios. Extensive experiments on 
 six datasets show that our \model significantly outperforms state-of-the-art methods, achieving average improvements of 12.25\% in AUC and 11.29\% in Macro-F1 across all datasets. 
\end{abstract}


\section{Introduction}
Anomaly detection~\cite{ding2021inductive,ma2021comprehensive,chai2022can}, aimed at identifying entities that deviate significantly from normal conditions, is extensively utilized across various applications, for example, fraudulent user detection in financial networks~\cite{huang2022dgraph,dou2020enhancing,liu2021intention,liu2021pick,chen2022antibenford}, anomalous behavior detection in social networks~\cite{yang2019mining,cheng2021causal,min2022divide,cao2024hierarchical}, malicious comments in review networks~\cite{tang2024dualgad}, review-scrubbing buffs in e-commerce networks~\cite{gong2023beyond,zhang2024dig}, and so on. Unlike time series anomaly detection, GAD presents greater challenges due to the inherent complexity of graph data~\cite{yang2023ahead,gao2023addressing,xu2024revisiting,zhang2022dual}, which typically encompasses node attributes and structural features~\cite{tang2022rethinking,zhou2023improving,liu2022dagad}. This complexity renders the task of capturing anomalies within graphs particularly formidable.

Due to the high cost of acquiring real anomaly data, a large number of existing GAD methods are performed in an unsupervised manner~\cite{xie2023unsupervised,xu2024unsupervised,zhang2023graph,wang2023unsupervised}, to detect instances that deviate significantly from the majority of data. Various methods have been proposed to solve the unsupervised graph anomaly detection (UGAD) problem. They can be broadly categorized into the following types: traditional methods, message passing-improved (MPI) methods, contrastive learning (CL)-based methods, graph autoencoder (GAE)-based methods, and Multi-view (MV) methods, etc. Early traditional UGAD methods such as~\cite{li2017radar} are usually based on machine learning algorithms to encode graph information and detect anomalies, or utilize residual information to capture anomalous features. Message passing-improved methods~\cite{luo2022comga,qiao2024truncated} learn anomalous node features by improving the message-passing mechanism of GNNs. Recently, with the rapid development of graph neural networks (GNNs), more and more CL-based~\cite{chen2022gccad} and GAE-based methods have emerged. For example, GRADATE~\cite{duan2023graph} performs anomaly detection through a multi-scale contrastive learning network with augmented views, and VGOD~\cite{huang2023unsupervised} combines variance-based and graph reconstruction-based models through contrastive learning to detect anomalies. ADA-GAD~\cite{he2024ada} builds a denoised graph first to train the graph encoder and then trains the graph decoder on the original graph for reconstructing node attributes and structure.

\textbf{Challenges.} Existing UGAD approaches, such as CL-based and GAE-based methods, have achieved promising results. Nevertheless, the UGAD task still faces the following two major challenges: 
(1) \textit{Most existing methods focus solely on non-multiplex heterogeneous graphs, while real-world graphs are often multiplex that typically include multiple types of interactions.} For example, there exist viewing, carting, and buying relations between users and items in e-commerce networks, or users' different comments and rating scores on items in review networks. These interactions result in complex structures, known as multiplex heterogeneous graphs~\cite{liu2020fast,yu2022multiplex,li2025dual}.  
Anomaly detection in multiplex heterogeneous graphs is challenging because the interacting multiple relations are extremely complex and different types of relations have different effects on anomaly detection.
(2) \textit{In real unsupervised scenarios, most of existing aspects face difficulties in selecting anomaly score thresholds because the number of anomalies is unknown}. Existing models typically employ two approaches to select the anomaly score threshold: 1) by selecting the threshold based on the known number of anomalies when ground truth information in the test set is available, and 2) by choosing the threshold that yields the best model performance. However, these methods are not suitable when the number of anomalies and their labels are unknown, making it unreasonable to rely on these approaches for threshold selection in the real unsupervised scenario with no ground truth information from the test set. 

\textbf{Presented Work.} Recognizing the above challenges, we focus on exploring the important impact of different interactive relations on node representation learning and extracting anomaly information through graph reconstructions for multiplex heterogeneous graphs. 
To this end, this paper proposes a novel unsupervised multiplex graph anomaly detection method, named \model. 
We first learn multi-relational correlations among nodes in multiplex heterogeneous graphs and capture anomaly information during node attribute and structure reconstruction in the original-view graph. Then, to further extract abnormal information, we generate attribute-level and subgraph-level augmented-view graphs respectively, and perform attribute and structure reconstruction in augmented graphs. Innovatively, we coordinate the fusion of GMAEs with multiple masking mechanisms across multiplex graphs to reconstruct node attributes, network structures, and subgraphs in the original and augmented graphs. Furthermore, we learn to optimize node attributes and structural features through contrastive learning between original-view and augmented-view graphs to improve the model's ability to detect anomalies. Meanwhile, we also propose a new anomaly score threshold selection strategy, which allows the model to be independent of the ground truth in real unsupervised scenarios. Extensive experiments on six real-world datasets with injected and real anomalies have demonstrated the effectiveness and efficiency of \model compared with state-of-the-art UGAD methods.


This work makes the following contributions:
\begin{itemize}[leftmargin=*]
    \item We propose a novel unsupervised multiplex graph anomaly detection method named UMGAD that emphasizes the importance of constructing and exploiting different interactions between nodes and addresses the UGAD problem in multiplex heterogeneous graphs. 
    \item We design an effective method for selecting appropriate anomaly score thresholds in the real unsupervised scenario without involving the ground truth by performing anomaly information extraction on both the original- and the augmented-view graphs. 
    \item We conduct extensive experiments on datasets with both injected and real anomalies to demonstrate the superiority and efficiency of our proposed method. Experiment results show that \model achieves 12.25\%, and 11.29\% average improvement in terms of AUC and Macro-F1 across six datasets compared to SOTA baselines.  
\end{itemize}

\section{Related Work}
UGAD has gained significant attention recently, with many researchers focusing on using self-supervised signals to enhance anomaly detection in unsupervised settings. Early methods like Radar~\cite{li2017radar} and ResGCN~\cite{pei2022resgcn} capture anomalies by analyzing the residuals of attribute information and its consistency with network data. With the rise of GNNs, UGAD models based on GNNs have emerged. We categorize GNN-based models into four types: MPI methods, CL-based methods, GAE-based methods, and MV methods.


\textbf{MPI methods.} Most GNN-based methods adopt message passing~\cite{wang2021modeling} to learn node representations by aggregating neighbor information. However, inconsistencies between normal and anomalous features can hinder detection. To address this, Luo et al.\cite{luo2022comga} enhance message passing via community segmentation and structure learning. Bei et al.\cite{bei2023reinforcement} employ reinforcement learning to expand neighbor pools, assess reliability, and amplify messages from trustworthy nodes. Qiao et al.~\cite{qiao2024truncated} propose TAM, which maximizes local affinity on a truncated graph to avoid full-graph noise.


\textbf{CL-based methods.} Contrastive learning, a popular approach for extracting self-supervised signals~\cite{liu2024towards,kong2024federated,liu2024bourne}, has been applied to UGAD tasks~\cite{wang2023cross} to capture feature inconsistencies between normal and anomalous nodes. CoLA~\cite{liu2021anomaly} samples instance pairs using local network information to embed high-dimensional attributes and structures. Zhang et al.\cite{zhang2022reconstruction} introduce Sub-CR, which combines multi-view contrastive learning with attribute reconstruction. GRADATE\cite{duan2023graph} extends contrastive learning to the subgraph level via node-subgraph and subgraph-subgraph comparisons. Huang et al.~\cite{huang2023unsupervised} integrate variance-based learning with attribute reconstruction for balanced anomaly detection.

\begin{figure*}[t]\label{framework}
    \begin{center}
    \includegraphics[width=1\textwidth]{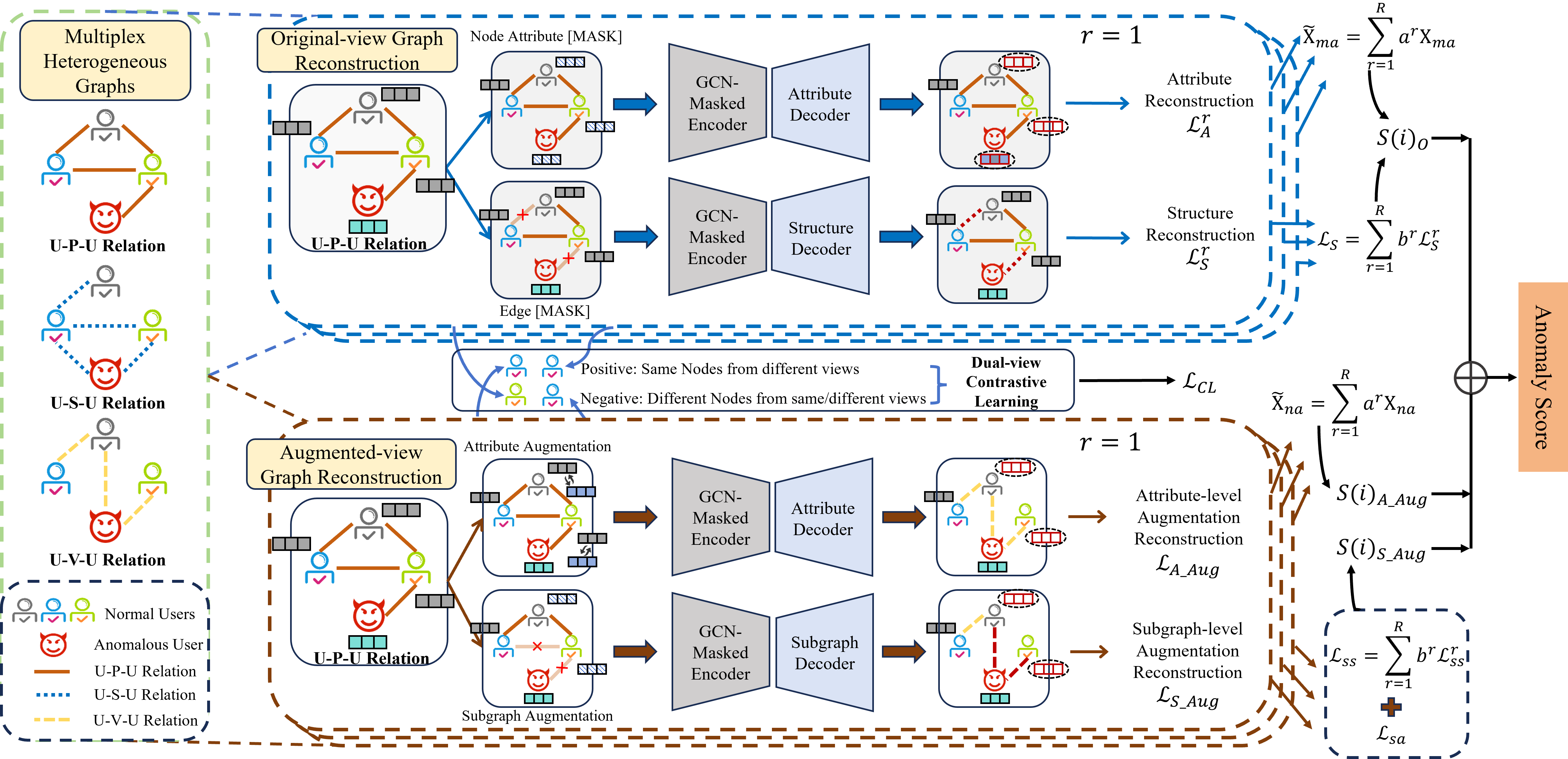}
    \caption{The overview of the proposed UMGAD. The multiplex heterogeneous graph on the left takes Amazon dataset as an example, where U-P-U links users reviewing at least one product, U-S-U links users having at least one identical star rating within a week, and U-V-U links linguistically similar users.}
    \label{fig:framework}
    \end{center}
    \vspace{-2mm}
\end{figure*}


\textbf{GAE-based methods.} GAE-based approaches~\cite{li2019specae} detect anomalies by reconstructing node attributes and structural features. Ding et al.\cite{ding2019deep} use GCNs to capture graph topology and node attributes for embedding learning. Roy et al.\cite{roy2024gad} propose GAD-NR, which reconstructs a node’s neighborhood—including structure, self-properties, and neighbor features—based on its embedding. ADA-GAD~\cite{he2024ada} introduces a two-stage denoising framework: the GAE is first trained on denoised graphs, then the encoder is retrained on original graphs with anomaly regularization to avoid overfitting.


\textbf{MV methods.} Recent work has explored anomaly detection on multiplex graphs, categorized as multi-view GAD methods. AnomMAN~\cite{chen2023anomman} detects anomalies in multi-view attribute networks by modeling cross-view attributes and interactions, using attention to fuse their importance. DualGAD~\cite{tang2024dualgad} employs a dual-bootstrapped self-supervised framework to address feature-structure inconsistency and imbalanced anomaly signals. It integrates subgraph reconstruction with cluster-guided contrastive learning, achieving state-of-the-art performance in node-level anomaly detection.

Our \model combines multi-view graph reconstruction with cross-view contrastive learning. Unlike single-view methods (e.g., GAE-based), it uses attribute-level and subgraph-level augmentations to capture richer relations. To avoid over-smoothing seen in MPI methods, \model employs masking and GMAE fusion for selective feature focus. It also introduces a novel, label-free thresholding strategy for unsupervised anomaly detection.
\section{Preliminary}
Generally, a network is denoted as $\mathcal{G} = \{\mathcal{V}, \mathcal{E}, \mathcal{X}\}$, where $\mathcal{V}$ is the collection of nodes, and $\mathcal{E}$ is the collection of edges between nodes, and $\mathcal{X}$ is the collection of node attributes. 

\begin{Def}[Multiplex Heterogeneous Graphs, MHGs]\label{def1}
Given the defined network $\mathcal{G}$, a multiplex heterogeneous graph can be divided into $R$ relational subgraphs $\mathcal{G}=\{\mathcal{G}^1, \mathcal{G}^2, \ldots, \mathcal{G}^R\}$, where $R$ denotes the number of interactive relation categories. Each subgraph is defined as $\mathcal{G}^r=\{\mathcal{V}, \mathcal{E}^r, \mathcal{X}\}$, where $\mathcal{V}$ and $\mathcal{X} \in \mathbb{R}^{|\mathcal{V}|\times f}$ denote the set of all nodes and node attributes respectively, $f$ is the size of the node attribute, $\mathcal{E}^r$ represents the edge set in the $r$-th relational subgraph, $\mathcal{R}=\{1,2,\ldots, R\}$ represents the set of edge relation categories, $|\mathcal{R}|=R$, and $\mathcal{E}=\bigcup\limits_{r\in \mathcal{R}} \mathcal{E}^r$ is the collection edge set of various relation subgraphs. 
\end{Def}

Next, we formally define our studied problem in this work. 

\begin{Pro*}[Unsupervised Multiplex Graph Anomaly Detection, UMGAD]
    Given a multiplex graph $\mathcal{G}$, the unsupervised GAD problem aims to identify nodes that significantly deviate from the majority in both structural features and node attributes. We try to define an anomaly function $S(i)$ that assigns an anomaly score to each node $v_i \in \mathcal{V}$. Nodes with scores exceeding the selected anomaly threshold are classified as anomalous, while others are considered normal. 
\end{Pro*}

\section{Methodology}
In this section, we propose a novel unsupervised multiplex graph anomaly detection method, named \model, depicted in Fig.~\ref{fig:framework}. It mainly includes three key components:  \textit{(1) Original-view Graph Reconstruction}, \textit{(2) Augmented-view Graph Reconstruction} and \textit{(3) Dual-view Contrastive Learning}. 

\subsection{Original-view Graph Reconstruction}
\subsubsection{Attribute Reconstruction}\label{AR}
Attribute inconsistency is one of the most important evidence to distinguish abnormal and normal nodes in the feature space. However, the aggregation mechanism of existing GNN models is based on the homogeneity assumption. It is detrimental to identifying rare abnormal nodes because most of the connected normal nodes will weaken the abnormal information during message propagation, thus smoothing the attribute inconsistency to hinder the abnormality detection, especially in MHGs. 
To learn the importance of different relations for UGAD in MHGs, we first decouple the original graph into multiple relational subgraphs. Then we mask each relational subgraph with an attribute masking strategy. Formally, we get the subgraph $\mathcal{G}^r$ and randomly sample a subset $\mathcal{V}_{ma}$ with a masking ratio $r_m$ and then obtain the perturbation subgraph as follows:
\begin{equation}\label{eq1}
    \mathcal{G}_{ma}^r = (\mathcal{V},\mathcal{E},\mathcal{X}_{ma}),
\end{equation}
where the original node attributes in $\mathcal{X}_{ma}$ are replaced by the [MASK] tokens, which are learnable vectors. Masked nodes are selected by utilizing uniform random sampling without replacement, which helps prevent potential bias. We repeat the attribute masking strategy on each relational subgraph for $K$ times to finally generate a set of masked subgraphs $\mathcal{G}_{ma}^r=\{\mathcal{G}_{ma}^{r,1},\mathcal{G}_{ma}^{r,2},\ldots,\mathcal{G}_{ma}^{r,K}\}$ for $\mathcal{G}^r$. These masked subgraphs are then fed into the GNN-based encoder and decoder, and the output of each encoder is:
\begin{equation}\label{eq2}
    \mathcal{X}_{ma}^{r,k}=Dec(Enc(\mathcal{V}_{re},\mathcal{E}^r|\mathbf{W}_{enc1}^{r,k}),\mathcal{E}^r|\mathbf{W}_{dec1}^{r,k}),
\end{equation}
where $\mathcal{X}_{ma}^{r,k} \in \mathbb{R}^{|\mathcal{V}|\times f}$ is the output of the attribute decoder in the $k$-th masking repeat of the $r$-th relational subgraph. $\mathcal{V}_{re}=\mathcal{V} \setminus \mathcal{V}_{ma}$ denotes the set of remaining nodes that are unaffected by masking, $\mathbf{W}_{enc1}^{r,k}\in \mathbb{R}^{f\times d_h}$ and $\mathbf{W}_{dec1}^{r,k} \in \mathbb{R}^{d_h\times f}$ are the trainable weight matrices, $d_h$ represents the dimension of latent vectors during graph convolution. Inside each encoder and decoder, before the vertical lines $|$ are the input variables, and after the vertical lines are the important learnable weights. In subsequent formulas, each vertical line means the same. 

After processing each decoupled relational subgraph, \model learns the importance (weights) of different relations collaboratively, instead of dealing with each relational subgraph independently. It fuses the effects of these relations on anomaly detection through a set of learnable weight parameters $a^r$, learnable weights as follows:
\begin{equation}\label{eq3}
    \tilde {\mathcal{X}}_{ma}^k = \sum\limits_{r = 1}^R {{a ^r} {\mathcal{X}_{ma}^{r,k}}},
\end{equation}
where $a^r$ is initially randomized using a normal distribution, and continuously optimized through the model's self-supervised training.

Finally, to optimize the attribute reconstruction, we compute the reconstruction error between the aggregated node attributes and the original attributes of the masked nodes as follows:
\begin{equation}\label{eq4}
    \mathcal{L}_A = {\sum\limits_{k = 1}^K}\frac{1}{{|{\mathcal{V}_{ma}^k}|}} {\sum\limits_{{v_i} \in {\mathcal{V}_{ma}^k}} {(1 - \frac{{\tilde x_{ma}^k(i)^\mathsf{T}{x(i)}}}{{||\tilde x_{ma}^k(i)|| \cdot ||x(i)||}})} ^\eta },\eta  \ge 1,
\end{equation}
which is the average reconstruction loss of all masked nodes. ${\mathcal{V}_{ma}^k}$ is the $k$-th masked node subset, $\tilde x_{ma}^k(i)$ denotes the $i$-th reconstructed node attribute vector in $\mathcal{X}_{ma}^{r,k}$, $x(i)$ denotes the $i$-th original node attribute vector in $\mathcal{X}$. The scaling factor $\eta$ is a hyperparameter that can be adjusted on different datasets. 

\subsubsection{Structure Reconstruction}\label{SR}
Besides attribute anomalies, structural anomalies are also more difficult to recognize, they can be camouflaged by mimicking the attributes of normal nodes. However, structural inconsistencies are reflected in the connections, and if the target node is not well reconstructed structurally, then it is likely to be anomalous.

Similar to node attribute reconstruction, we capitalize on the inconsistency of the structure space and introduce a structural (edge) masking mechanism that works to break short-range connections, causing nodes to look elsewhere for evidence that suits them. We set masking ratios as $r_m$ to randomly sample an edge subset $\mathcal{E}_{ms}^r$ from the edges observed in the subgraph and then obtain the corresponding perturbed subgraph as follows:
\begin{equation}\label{eq5}
    \mathcal{G}_{ms}^r = (\mathcal{V},\mathcal{E}^r \setminus \mathcal{E}_{ms}^r,\mathcal{X}),
\end{equation}
where $\mathcal{E}^r \setminus \mathcal{E}_{ms}^r$ denotes the remaining edges set after edge masking in the $r$-th subgraph. Similarly, We utilize the same random sampling method as attribute masking to select the masked edges. We repeat the structure masking strategy on all relational subgraphs for $K$ times to finally generate a set of masked subgraphs $\mathcal{G}_{ms}^r=\{\mathcal{G}_{ms}^{r,1},\mathcal{G}_{ms}^{r,2},\ldots,\mathcal{G}_{ms}^{r,K}\}$ for each relational subgraph $\mathcal{G}^r$, which are further fed into the GNN-based encoder and decoder to learn the node attribute:
\begin{equation}\label{eq6}
    \mathcal{X}_{ms}^{r,k}=Dec(Enc(\mathcal{V},\mathcal{E}^r \setminus \mathcal{E}_{ms}^{r,k}|\mathbf{W}_{enc2}^{r,k}),\mathcal{E}^r \setminus \mathcal{E}_{ms}^{r,k}|\mathbf{W}_{dec2}^{r,k}),
\end{equation}
where $\mathcal{X}_{ms}^{r,k} \in \mathbb{R}^{|\mathcal{V}|\times f}$ is the output of the attribute decoder in the $k$-th masking repeat of the $r$-th relational subgraph, $\mathbf{W}_{enc2}^{r,k} \in \mathbb{R}^{f\times d_h}$ and $\mathbf{W}_{dec2}^{r,k} \in \mathbb{R}^{d_h\times f}$ are the trainable weight matrices. In contrast to the reconstruction goal of attribute masking, we use the edge set of the $k$-th masked sampling subgraph of the $r$-th relational subgraph, $\mathcal{E}_{ms}^{r,k}$, as a self-supervised signal to recover the original subgraph structure by predicting the masked edges with the cross-entropy function as follows:
\begin{equation}\label{eq7}
    \mathcal{L}_S^r = \sum\limits_{k=1}^K \sum\limits_{(v,u) \in {\mathcal{E}_{ms}^{r,k}}} {\log \frac{{\exp (g(v,u))}}{{\sum\nolimits_{(v,u' \in {\mathcal{E}_{ms}^{r,k}})} {\exp (g(v,u'))} }}},
\end{equation}
where $g(v,u)={x_{ms}^{r,k}(v)}^{\mathsf{T}}x_{ms}^{r,k}(u)$ calculates the estimated probability of link between node $v$ and $u$, and then we can obtain the reconstructed adjacent matrix $\tilde{\mathbf{A}}_{O}^{r}$. We introduce negative sampling to train a more generalized model by accumulating all edge prediction probabilities of the unmasked subgraphs in the denominator, making the model robust to external noise. Notably, the standard normalization in Eq.~\eqref{eq4} is used to calculate the similarity between the reconstructed attribute features and the original attribute features. The softmax normalization in Eq.~\eqref{eq7} is used to evaluate the probability of a link being reconstructed between two nodes.

Similar to Eq.~\eqref{eq3}, collaboratively considering the importance of different relational subgraphs for structure reconstruction, we aggregate all subgraph structure reconstruction losses together using a set of learnable weight parameters $b^r$ as follows:
\begin{equation}\label{eq8}
    \mathcal{L}_S = \sum\limits_{r = 1}^R {{b ^r}{\mathcal{L}_S^r}},
\end{equation}
where $b^r$ is initially randomized and continuously optimized through the model's self-supervised training, similar to $a^r$.

The learned embedding captures anomalies in the attribute and structure space by jointly masking the attributes and edge reconstruction of the subgraph. Finally, the training objective for the original-view graph reconstruction is as follows:
\begin{equation}\label{eq9}
    \mathcal{L}_{O}=\alpha \mathcal{L}_A + (1-\alpha) \mathcal{L}_S, 
\end{equation}
where $\alpha \in (0,1)$ is the hyperparameter to balance the importance between attribute reconstruction
loss $\mathcal{L}_A$ and structure reconstruction loss $\mathcal{L}_S$.

\subsection{Augmented-view Graph Reconstruction}
Due to the small number of anomalies, graphs containing anomalous nodes are usually unbalanced, which may disturb the detection of anomalous nodes. Therefore, we introduce two simplified graph masking strategies to generate two levels of augmented graphs, namely: attribute-level augmented graph and subgraph-level augmented graph, to reduce the redundant information in the original graph. Note that in the augmented-view graph reconstruction module, we do not actually adopt traditional graph data augmentation methods. Instead, we perform attribute-level augmentation by swapping the attributes of different nodes and subgraph-level augmentation through subgraph masking, without introducing new content.

\subsubsection{Attribute-level Augmented Graph Reconstruction}\label{AAR}
For this module, we randomly select a subset of nodes $\mathcal{V}_{na} \subset \mathcal{V}$ for replacement-based augmentation. The selected node features are adjusted as follows:
\begin{equation}\label{eq10}
    \tilde{x}(i) = \begin{cases} 
x(j), &  i \in \mathcal{V}_{aa} \\
x(i), &  i \in \mathcal{V} \setminus \mathcal{V}_{aa} 
\end{cases}.
\end{equation}

We randomly select another node $j$ and replace the original feature $x(i)$ of $i$ with the feature $x(j)$ of node $j$ if $i \in \mathcal{V}_{aa}$. Based on this strategy, we obtain the attribute-level augmented graph, $\mathcal{G}_{aa}$. Then we introduce the masking mechanism that only masks the augmented node set $\mathcal{V}_{aa}$, where the feature of each node $i \in \mathcal{V}_{aa}$ is masked.

We repeat the above augmentation operation for $K$ times to generate a collection of the attribute-level augmented graphs for each relational subgraph $\mathcal{G}^r$, denoted as $\mathcal{G}_{aa}^r = \{\mathcal{G}_{aa}^{r,1}, \mathcal{G}_{aa}^{r,2},\ldots, \mathcal{G}_{aa}^{r,K}\}$, where each $\mathcal{G}_{aa}^{r,k} = (\mathcal{V}, \mathcal{E}^r, \tilde{\mathcal{X}}_{aa}^{r,k})$, and $\tilde{\mathcal{X}}_{aa}^{r,k}$ is the augmented attribute matrix generated each time. We feed each $\mathcal{G}_{aa}^{r,k}$ into the attribute-level GMAE consisting of the simplified GCN encoder and decoder with a masking mechanism. Then we can obtain the reconstructed node embedding matrix $\mathcal{X}_{na}^{r,k}$ as follows:
\begin{equation}\label{eq11}
    \mathcal{X}_{aa}^{r,k}=Dec(Enc(\mathcal{V},\mathcal{E}^r|\mathbf{W}_{enc3}^{r,k}),\mathcal{E}^r|\mathbf{W}_{enc3}^{r,k}). 
\end{equation}

Similarly, we aggregate all subgraph attributes by using the set of learnable weights $a^r$ as follows:
\begin{equation}\label{eq12}
    \tilde{\mathcal{X}}_{aa}^k = \sum\limits_{r = 1}^R {{a ^r}{\mathcal{X}_{aa}^{r,k}}}. 
\end{equation}

Therefore, the training objective for the attribute-level augmented-view graph reconstruction is defined as follows:
\begin{equation}\label{eq13}
    \mathcal{L}_{A\_Aug}= \frac{1}{{|{\mathcal{V}_{aa}^k}|}}{\sum\limits_{k=1}^K}{\sum\limits_{{v_i} \in {\mathcal{V}_{aa}^k}} {(1 - \frac{{\tilde{x}_{aa}^k(i)^\mathsf{T}{x(i)}}}{{||\tilde{x}_{aa}^k(i)|| \cdot ||x(i)||}})} ^\eta },\eta  \ge 1,
\end{equation}
where $|{\mathcal{V}_{aa}^k}|$ is the $k$-th masked subset, $\tilde x_{aa}^k(i)$ denotes the reconstructed node attribute vector in $\mathcal{X}_{aa}^{k}$, $x(i)$ denotes the original node attribute vector in $\mathcal{X}$. 

\subsubsection{Subgraph-level Augmented Graph Reconstruction}\label{SAR}
First, we propose a subgraph masking mechanism that employs a subgraph sampling strategy based on random walk with restart for subgraph sampling and masks these sampled subgraphs for each relational subgraph $\mathcal{G}^r$ in $\mathcal{G}$. Then we obtain the subgraph-level augmented graphs collection $\mathcal{G}_{s}^r = \{\mathcal{G}_{s}^{r,1}, \mathcal{G}_{s}^{r,2},\ldots, \mathcal{G}_{s}^{r,K}\}$, where $\mathcal{G}_{s}^{r,k} = (\mathcal{V}, \mathcal{E}^r \setminus \mathcal{E}_s^r, \mathcal{X}_{s}^{r,k})$, $\mathcal{E}_s^r$ and $\mathcal{X}_{s}^{r,k}$ are the edge subset and generated attribute matrix.

Subgraph-level augmentation can be considered as a specific combination of node attribute level and structure level augmentation. The reconstructed node attribute matrix $\tilde{\mathcal{X}}_{sa}^k$ and structural adjacency matrix $\tilde{\mathbf{A}}_{ss}^{r}$ are shown below, respectively:
\begin{equation}\label{eq14}
    \tilde{\mathcal{X}}_{sa}^k = \sum\limits_{r = 1}^R {{a ^r}{\mathcal{X}_{sa}^{r,k}}},\tilde{\mathbf{A}}_{ss}^{r} = \sum\limits_{k = 1}^K \sigma((\tilde{\mathcal{X}}_{sa}^k)^{\mathsf{T}}{\tilde{\mathcal{X}}_{sa}^k}). 
\end{equation}

Then the subgraph-level augmented attribute and structure reconstruction loss values $\mathcal{L}_{sa}$ and $\mathcal{L}_{ss}$ are defined as follows:
\begin{align}\label{eq15}
    \mathcal{L}_{sa} = \frac{1}{{|{\mathcal{V}_{sa}^k}|}}{\sum\limits_{k=1}^K}{\sum\limits_{{v_i} \in {\mathcal{V}_s^k}} {(1 - \frac{{\tilde{x}_{sa}^k(i)^\mathsf{T}{x(i)}}}{{||\tilde{x}_{sa}^k(i)|| \cdot ||x(i)||}})} ^\eta },\eta  \ge 1, \notag \\
    \mathcal{L}_{ss} = \sum\limits_{r = 1}^R {b^r} \sum\limits_{k=1}^K {\sum\limits_{(v,u) \in {\mathcal{E}_{ss}^{r,k}}} {\log \frac{{\exp (g(v,u))}}{{\sum\nolimits_{(v,u' \in {\mathcal{E}_{ss}^{r,k}})} {\exp (g(v,u'))} }}} },  
\end{align}
where $|{\mathcal{V}_{sa}^k}|$ is the $k$-th masked subset, $\tilde{x}_{sa}^k(i)$ and $x(i)$ are the attribute-level augmented embedding and original node attribute respectively, and $g(v,u)={x_s^{r,k}(v)}^{\mathsf{T}}x_s^{r,k}(u)$ denotes the estimated probability of the link between node $v$ and $u$.

Finally, the training objective for the subgraph-level augmented-view graph reconstruction is defined as follows:
\begin{equation}\label{eq16}
    \mathcal{L}_{S\_Aug}=\beta \mathcal{L}_{sa} + (1-\beta) \mathcal{L}_{ss}.
\end{equation}
where $\beta \in (0,1)$ is the hyperparameter to balance the importance between attribute and structure reconstruction loss, $\mathcal{L}_{sa}$ and $\mathcal{L}_{ss}$, in the subgraph-level augmentation.

It is worthwhile to emphasize that normal nodes tend to have more similar features than their neighbors, while anomalous nodes exhibit greater feature differences. Through masking and GNN-based reconstruction, normal nodes are more likely to be accurately reconstructed, whereas anomalous nodes are less likely to be. Therefore, we improve the model's ability to identify anomalies by employing different masking mechanisms with a graph autoencoder.

\subsection{Dual-view Contrastive Learning}
Dual-view contrastive learning is defined between the original view and the augmented view to learn more representative and intrinsic node embeddings, which helps capture further anomaly information. Specifically, we optimize the node attribute under joint loss by comparing the original-view graph with the attribute-level augmented-view graph and the subgraph-level augmented-view graph, respectively. The node attribute of node $i$ in the original view forms a positive pair with the augmented view and a negative pair with the attribute of node $j$ in the original and augmented views. We use the following loss function to optimize contrastive learning:
\begin{align}\label{eq17}
    \mathcal{L}_{cl}^{oa} = - \sum\limits_{i=1}^{|\mathcal{V}|} \log \frac{{e^{{{\tilde x}_{ma}}(i) \cdot {{\tilde x}_{aa}}(i)}}}{{e^{{{\tilde x}_{ma}}(i) \cdot {{\tilde x}_{ma}}(j)} + e^{{{\tilde x}_{ma}}(i) \cdot {{\tilde x}_{aa}}(j)}}}, \notag \\
    \mathcal{L}_{cl}^{os} = - \sum\limits_{i=1}^{|\mathcal{V}|} \log \frac{{e^{{{\tilde x}_{ma}}(i) \cdot {{\tilde x}_{sa}}(i)}}}{{e^{{{\tilde x}_{ma}}(i) \cdot {{\tilde x}_{ma}}(j)} + e^{{{\tilde x}_{ma}}(i) \cdot {{\tilde x}_{sa}}(j)}}},
\end{align}
where $\mathcal{L}_{cl}^{oa}$ represents the contrastive loss between the original-view graph and the attribute-level augmented-view graph, $\mathcal{L}_{cl}^{os}$ represents the contrastive loss between the original-view graph and the subgraph-level augmented-view graph. The final loss dual-view contrastive learning is $\mathcal{L}_{CL}=\mathcal{L}_{cl}^{oa}+\mathcal{L}_{cl}^{os}$.

\subsection{Optimization Objective}
Putting it all together, we have the
The overall loss function in the training stage is as follows:
\begin{equation}\label{eq18}
    \mathcal{L} = \mathcal{L}_{O} + \lambda \mathcal{L}_{A\_Aug} + \mu \mathcal{L}_{S\_Aug} + \Theta \mathcal{L}_{CL},
\end{equation}
where $\lambda$, $\mu$, and $\Theta$ are hyperparameters that measure the importance of diverse augmented views and contrastive learning.

Based on the final training loss $\mathcal{L}$, we train several autoencoder modules, sort the nodes based on their anomaly scores, $S(i)$, and treat the nodes with higher $S(i)$ as anomalous nodes. For both original-view and augmented-view graphs, we use the following formula to calculate and finally take the arithmetic average of multiple views as the anomaly score of node $v_i$:
\begin{equation}\label{eq19}
    S(i)_{*} = \varepsilon \cdot \left\| \tilde{x}_{*}(i) - x(i) \right\|_1 
    + (1 - \varepsilon) \cdot \frac{1}{R} \sum\limits_{r=1}^R \left\| \tilde{\zeta}_{*}^{r}(i) - \zeta^r(i) \right\|_2,
\end{equation}
where ${*}$ denotes the specific original $(O)$, attribute-level augmented $(A\_Aug)$, or subgraph-level augmented $(S\_Aug)$ graph, $|| \cdot |{|_1}$ denotes the Euclidean norm of a vector, $|| \cdot |{|_2}$ denotes the L1-norm of the vector, $\tilde{\zeta}_{*}^{r}(i)$ and $\zeta^r(i)$ represent the $i$-th row of the specific reconstructed structure matrix $\tilde{\mathbf{A}}_{*}^{r}$ and original structure matrix $\mathbf{A}^r$, respectively. Similarly, $\tilde{x}_{*}(i)$ and $x(i)$ are the reconstructed attribute vector and original attribute vector of the node $i$, respectively. Therefore, the anomaly score $S(i)$ of node $i$ is the mean value of all $S(i)_{*}$.

\subsection{Unsupervised Anomaly Score Threshold Selection Strategy}
For the unsupervised graph anomaly detection task, since the nodes' labels (normal or abnormal) are unknown, a common approach is to compute the anomaly score for each node and rank them in order of the anomaly score from high to low. Nodes with high anomaly scores are considered anomalous nodes. However, determining a more accurate anomaly score threshold has always been a difficult problem in UGAD. 

To determine an anomaly score threshold for unsupervised graph anomaly detection, we propose a strategy that combines moving average smoothing and inflection point detection. This approach works directly on the ordered anomaly score sequence, without relying on ground truth information from the test set (e.g., the number of anomalous nodes), making it suitable for real unsupervised scenarios.

Given the learned anomaly score set for $|\mathcal{V}|$ nodes, we first sort the scores in descending order to form $S = \{s(1), s(2), \dots, s(|\mathcal{V}|) \} $, where $s(1) \geq s(2) \geq \dots \geq s(|\mathcal{V}|)$. The sorted sequence reflects the relative magnitude of anomaly scores, with higher scores more likely corresponding to anomalous nodes. To mitigate the effects of noise and ensure a smooth transition in the sequence, we apply a moving average smoothing technique. For a given sliding window size $w$, the smoothed score sequence $\bar{S}$ is computed as:
\begin{equation}\label{eq20}
    \bar{s}(i) = \frac{1}{w} \sum_{j=i}^{i+w-1} s(j), \quad i = 1, 2, \dots, |\mathcal{V}|-w+1.
\end{equation}

The choice of $w$ balances noise reduction and the preservation of meaningful variations in the scores. A practical guideline is to set $w = \max(\lfloor 0.0001|\mathcal{V}| \rfloor, 5)$, which works well across different data scales. After smoothing, we calculate the first-order and second-order differences of the sequence to analyze its structural changes. The first-order difference is:
\begin{equation}\label{eq21}
    \Delta_1(i) = \bar{s}(i) - \bar{s}(i+1), \quad i = 1, 2, \dots, |\mathcal{V}|-w.
\end{equation}

While the second-order difference is defined as follows:
\begin{equation}\label{eq22}
    \Delta_2(i) = \Delta_1(i) - \Delta_1(i+1), \quad i = 1, 2, \dots, |\mathcal{V}|-w-1.
\end{equation}

Here, $\Delta_1(i)$ captures the local decrease in scores, and $\Delta_2(i)$ identifies changes in the rate of decrease. The second-order difference captures the inflection point where the decline in anomaly scores transitions from steep (anomalous nodes) to stable (normal nodes). The inflection point $k$ is identified as the index $i$ where the magnitude of $\Delta_2(i)$ reaches its maximum:
\begin{equation}\label{eq23}
    \mathcal{T} = \arg\max_i |\Delta_2(i)|, \quad i = 1, 2, \dots, |\mathcal{V}|-w-1,
\end{equation}
which corresponds to the region where the anomaly score sequence transitions from a steep decline to a plateau. The anomaly score threshold $s(\mathcal{T})$ is then defined as the smoothed score at the inflection point $\mathcal{T}$. If there exist several selectable points consistent with Eq.~\eqref{eq23}, we select the one with the smallest difference from $\bar{s}(|\mathcal{V}|)$ as $\mathcal{T}$. Therefore, nodes with scores $s(i) \geq s(\mathcal{T})$ are considered anomalous, while those with scores $s(i) < s(\mathcal{T})$ are deemed normal.

We believe that a good anomaly detection model should be able to clearly distinguish between anomalous and normal nodes based on anomaly scores. In other words, the anomaly scores, when sorted in descending order, should gradually stabilize and show a clear inflection point. Before the inflection point (with relatively higher anomaly scores), the nodes are likely to be anomalous, while after the inflection point (with smaller and less differentiated scores), the nodes should be considered normal. This method leverages the distribution of anomaly scores to identify a natural threshold, without requiring any ground truth labels. The combination of smoothing and inflection point detection ensures the approach is robust to noise while effectively distinguishing between anomalous and normal nodes.

\subsection{Time Complexity Analysis}
The proposed \model consists of three critical components: original-view graph reconstruction, augmented-view graph reconstruction, and dual-view contrastive learning.
In original-view graph reconstruction, the time complexity of GMAE is $O(|\mathcal{V}| \times L \times f)$, and the time complexity of attention aggregation operation is $O(|\mathcal{V}| \times d_h \times f \times R)$, where $|\mathcal{V}|$ denotes the number of nodes, and $L$ represents the number of Simplified GCN layers, $d_h$ and $f$ denote the dimension of latent vectors during graph convolution and the size of node attribute, $R$ is the number of relational subgraphs. The overall time complexity of original-view graph reconstruction is $O(|\mathcal{V}| \times f \times (L + d_h \times R)$. 
For augmented-view graph reconstruction, the overall time complexity is $O(|\mathcal{V}| \times f \times (L + d_h \times R)$.
The time complexity for dual-view contrastive learning is $O(|\mathcal{V}|\times f^2 + |\mathcal{V}|\times f)$. Hence, the entire time complexity of the proposed \model is approximately $O(|\mathcal{V}| \times f \times (L + d_h \times R + f)$. 
\section{Experiments}
In this section, we evaluate the performance of our proposed method through extensive experiments and answer the following questions: 
\begin{itemize}[leftmargin=*]
    \item \textbf{(RQ1)} How does \model effectively alleviate the difficulty of selecting anomaly score thresholds in UGAD? 
    \item \textbf{(RQ2)} How does \model perform compared to other GAD methods in real unsupervised scenarios? 
    \item \textbf{(RQ3)} How does \model perform on large-scale graphs?
    \item \textbf{(RQ4)} What are the effects of different modules in \model on performance? 
    \item \textbf{(RQ5)} How do different hyperparameter settings affect the performance? 
    \item \textbf{(RQ6)} How does \model perform compared to other GAD methods in the scenario of threshold selection with ground truth leakage?
    \item \textbf{(RQ7)} How does UMGAD perform in terms of time efficiency compared to SOTA methods?
\end{itemize}

\subsection{Experimental Settings}
\subsubsection{Dataset}\label{sec.datasets} We have conducted experiments on two datasets with injected anomalies: Retail\_Rocket~\cite{ren2024sslrec} (Retail for short) and Alibaba~\cite{fu2023multiplex}, and four real-world publicly available datasets with anomalies: Amazon~\cite{mcauley2013amateurs}, YelpChi~\cite{rayana2015collective}, DGraph-Fin (DG-Fin for short)~\cite{tang2023gadbench} and T-Social~\cite{tang2022rethinking}. The detailed description of these datasets is shown as follows. 
\begin{itemize}[leftmargin=*]
    \item \textbf{Retail\_Rocket (Retail for short)} contains user-item interactions such as page views (View), add-to-cart (Cart), and transactions (Buy).
    \item \textbf{Alibaba} includes three types of user-item interactions like Retail: View, Cart, and Buy.
    \item \textbf{Amazon} contains three types of user interactions: U-P-U links users who have reviewed at least one common product, U-S-U links users with identical star ratings within a week, and U-V-U links linguistically similar users.
    \item \textbf{YelpChi} contains three interactive relations among users: R-U-R connects the reviews posted by the same user, R-S-R connects reviews that have the same identical star rating, R-T-R connects reviews that posted in the same term.
    \item \textbf{DGraph-Fin (DG-Fin for short)} contains three interactive relations among users: U-C-U connects the emergency contact among users, U-B-U connects the borrowing behavior among users, U-R-U indicates that users follow each other.
    \item \textbf{T-Social} contains three interactive relations among users: U-R-U links users who are friends with each other, U-F-U links users having fraud or money laundering behavior, U-G-U links online gambling users.
\end{itemize}
\begin{table}[t]
\centering
\footnotesize
\caption{Statistical information of evaluation datasets. \#Ano. denotes the number of anomalies, (I) represents the injected anomalies, and (R) represents the real anomalies.}
\begin{tabular}{@{}>{\centering\arraybackslash}p{1.2cm}|>{\centering\arraybackslash}p{1.2cm}>{\centering\arraybackslash}p{1.5cm}>{\centering\arraybackslash}p{1.8cm}>{\centering\arraybackslash}p{1.5cm}@{}}
\toprule
    Datasets & \#Nodes & \#Ano. (I/R) & Relation Type & \#Edges \\
\midrule
    Retail & 32,287 & 300 (I) & \makecell[c]{View \\ Cart \\ Buy} & \makecell[c]{75,374 \\ 12,456 \\ 9,551}  \\
\midrule
    Alibaba & 22,649 & 300 (I) & \makecell[c]{View \\ Cart \\ Buy} & \makecell[c]{34,933 \\ 6,230 \\ 4,571}  \\
\midrule
    Amazon & 11,944 & 821 (R) & \makecell[c]{U-P-U \\ U-S-U \\ U-V-U} & \makecell[c]{175,608 \\ 3,566,479 \\ 1,036,737}  \\
\midrule
    YelpChi & 45,954 & 6,674 (R) & \makecell[c]{R-U-R \\ R-S-R \\ R-T-R} & \makecell[c]{49,315 \\ 3,402,743 \\ 573,616}  \\
\midrule
    DG-Fin & 3,700,550 & 15,509 (R) & \makecell[c]{U-C-U \\ U-B-U \\ U-R-U} & \makecell[c]{441,128 \\ 2,474,949 \\ 1,384,922}  \\
\midrule
    T-Social & 5,781,065 & 174,010 (R) & \makecell[c]{U-R-U \\ U-F-U \\ U-G-U} & \makecell[c]{67,732,284 \\ 3,025,679 \\ 2,347,545}  \\
\bottomrule
\end{tabular}
\label{tab:datasets}
\end{table}

For anomaly injection, we follow the method in~\cite{ding2019interactive} to introduce both structural and attribute anomalies. For structural anomalies, a clique of size $m$ is formed by randomly selecting $m$ nodes and fully connecting them with one or multiple randomly assigned relation types. All $m$ nodes in the clique are considered anomalies. This process is repeated iteratively until $n$ factions are created, yielding $m \times n$ structural anomalies. For attribute anomalies, we randomly select $m \times n$ nodes and for each, identify the most deviant attribute by maximizing the Euclidean distance ${||x_i - x_j||}^2$ between node $i$ and a randomly selected node $j$. We then update node $i$’s attribute to match node $j$'s. For datasets with real anomalies, we directly use publicly available datasets.

\subsubsection{Baselines} We mainly compare five categories of methods on the unsupervised GAD task. 
\begin{figure*}[t]
    \centering
    \begin{subfigure}{0.34\textwidth}
        \includegraphics[width=\linewidth]{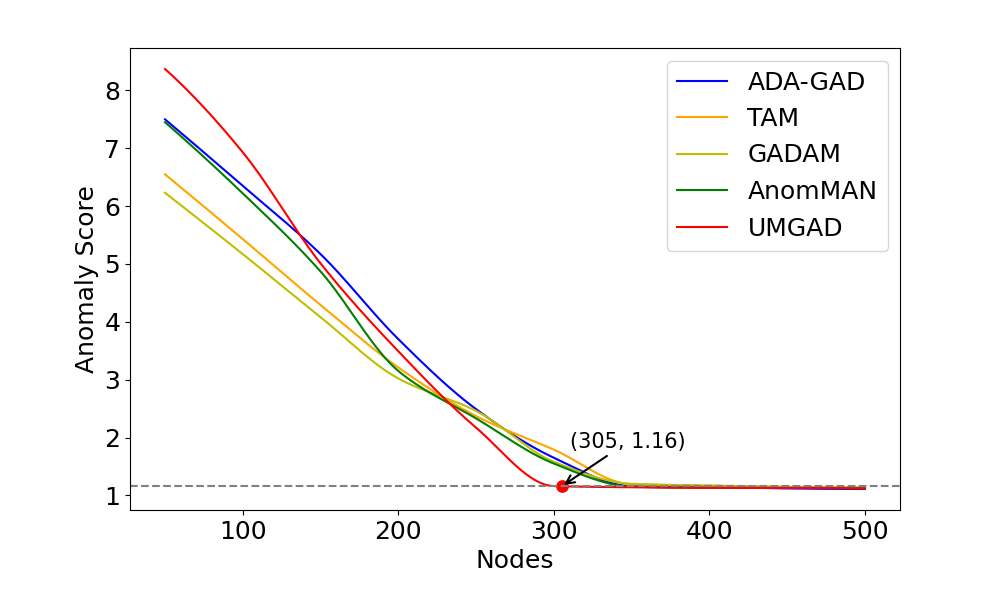}
        \vspace{-6mm}
        \caption{Retail (\#injected anomalies: 300)}
        \label{fig:threshold_Retail}
    \end{subfigure}
    \hspace{-6mm}
    \begin{subfigure}{0.34\textwidth}
        \includegraphics[width=\linewidth]{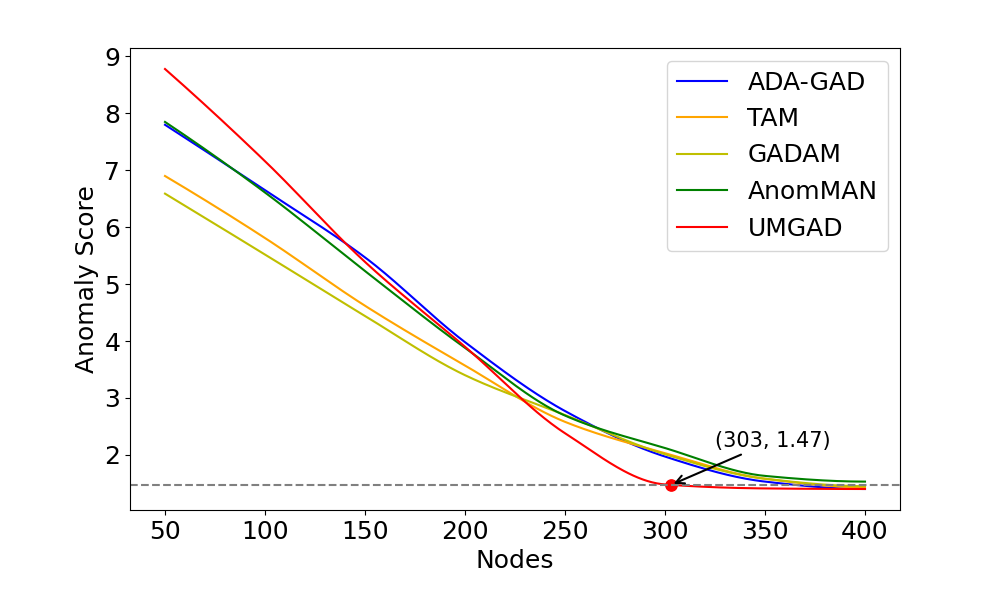}
        \vspace{-6mm}
        \caption{Alibaba (\#injected anomalies: 300)}
        \label{fig:threshold_Alibaba}
    \end{subfigure}
    \hspace{-6mm}
    \begin{subfigure}{0.34\textwidth}
        \includegraphics[width=\linewidth]{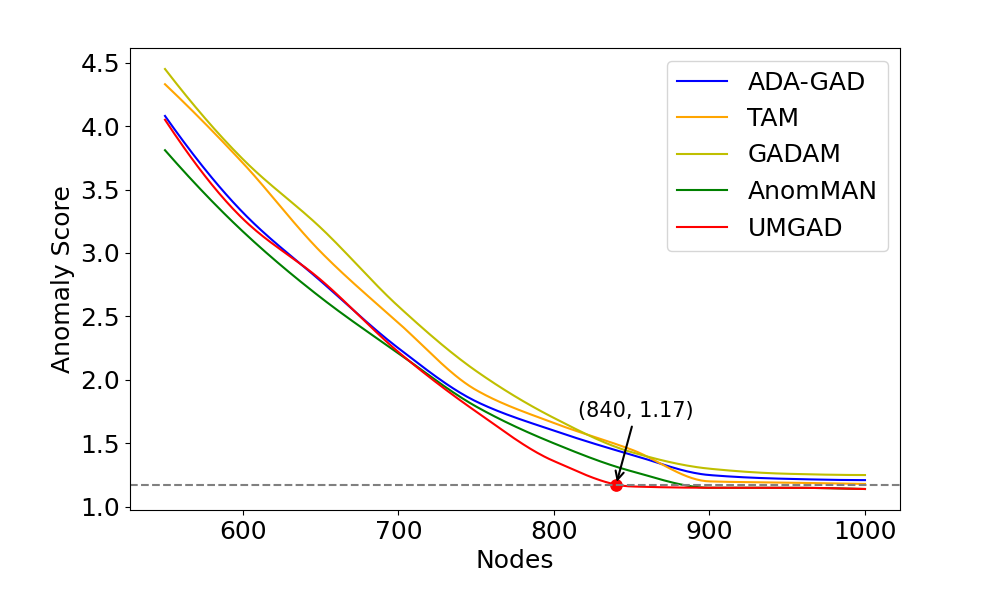}
        \vspace{-6mm}
        \caption{Amazon (\#real anomalies: 821)}
        \label{fig:threshold_Amazon}
    \end{subfigure}
    \begin{subfigure}{0.34\textwidth}
        \includegraphics[width=\linewidth]{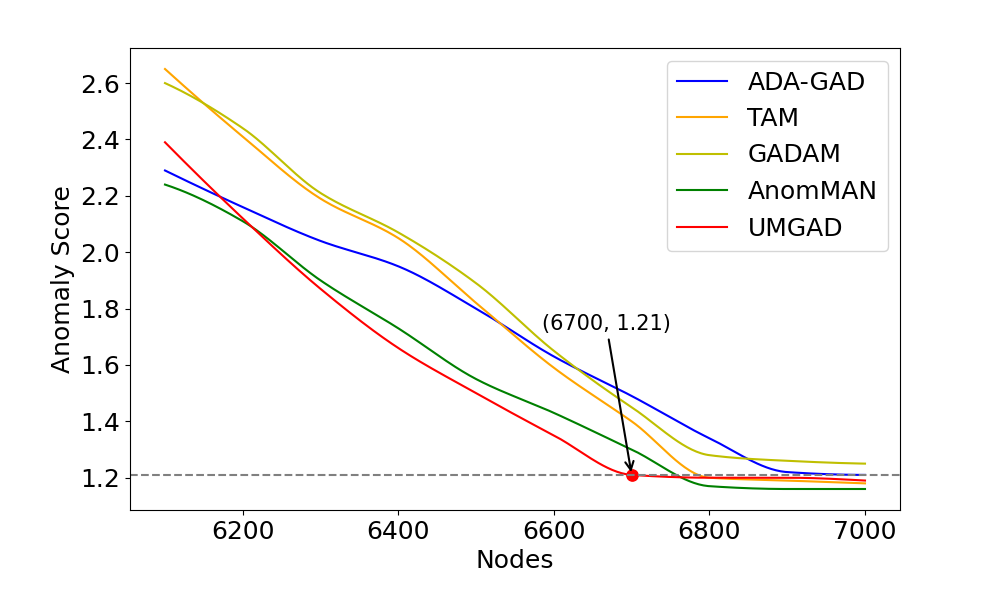}
        \vspace{-6mm}
        \caption{YelpChi (\#real anomalies: 6,674)}
        \label{fig:threshold_YelpChi}
    \end{subfigure}
    \hspace{-6mm}
    \begin{subfigure}{0.34\textwidth}
        \includegraphics[width=\linewidth]{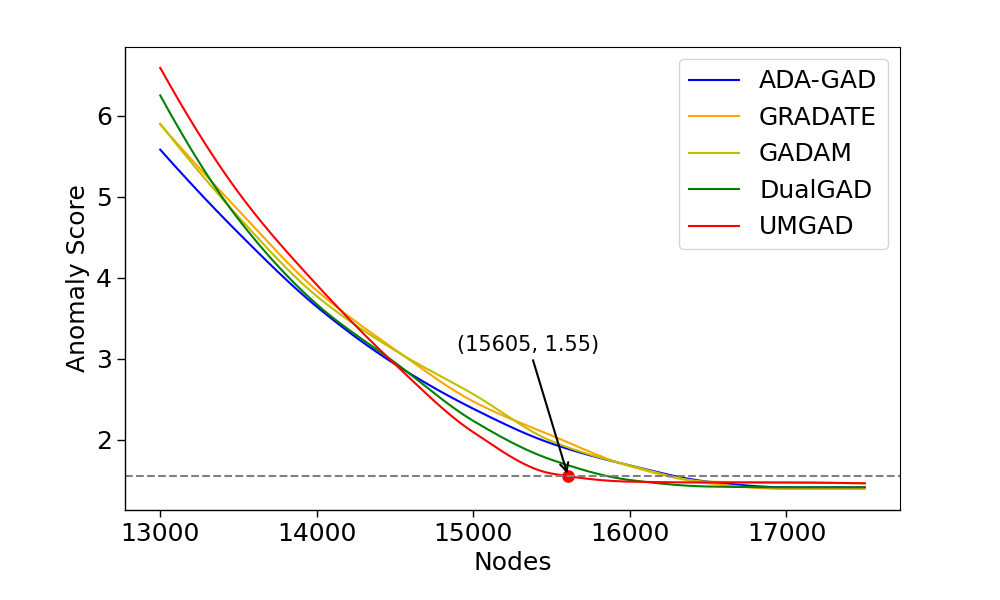}
        \vspace{-6mm}
        \caption{DG-Fin (\#real anomalies: 15,509)}
        \label{fig:threshold_DGFin}
    \end{subfigure}
    \hspace{-6mm}
    \begin{subfigure}{0.34\textwidth}
        \includegraphics[width=\linewidth]{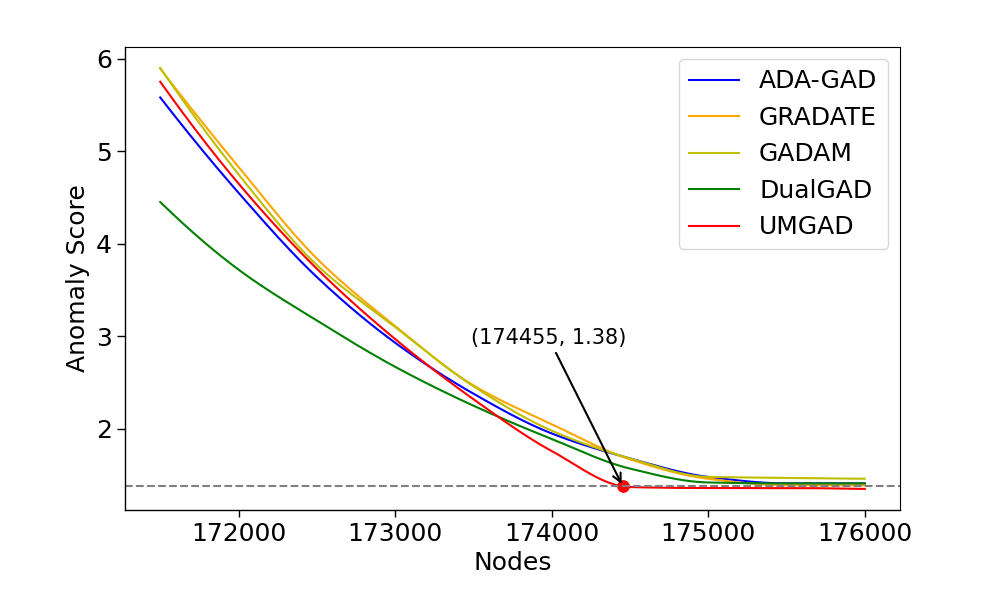}
        \vspace{-6mm}
        \caption{T-Social (\#real anomalies: 174,010)}
        \label{fig:threshold_TSocial}
    \end{subfigure}
    \caption{Visualization of ranked node anomaly scores for SOTA methods on four datasets.}
    \label{fig:threshold}
    \vspace{-4mm}
\end{figure*}

\noindent
\textbf{Traditional GAD Method:}
\begin{itemize}[leftmargin=*]
    \item \textbf{Radar~\cite{li2017radar}} characterizes the residuals of attribute information and its consistency with network information.
\end{itemize}
\textbf{Specific MPI Methods:}
\begin{itemize}[leftmargin=*]
    \item \textbf{ComGA~\cite{luo2022comga}} enhances GNNs through community-based segmentation and feature learning.
    \item \textbf{RAND~\cite{bei2023reinforcement}} integrates reinforced neighborhood selection and improves the message-passing mechanism for UGAD.
    \item \textbf{TAM~\cite{qiao2024truncated}} learns embeddings by maximizing local affinity and removing non-homogeneous edges to reduce bias.
\end{itemize}
\textbf{CL-based Methods:}
\begin{itemize}[leftmargin=*]
    \item \textbf{CoLA~\cite{liu2021anomaly}} 
    proposes a GNN-based contrastive model that captures high-dimensional attributes and local structures.
    \item \textbf{ANEMONE~\cite{jin2021anemone}} combines a multi-scale contrastive GNN encoder and a statistical consistency estimator.
    \item \textbf{Sub-CR~\cite{zhang2022reconstruction}} combines multi-view comparison and attribute reconstruction to detect anomalies on attribute networks.
    \item \textbf{ARISE~\cite{duan2023arise}} identifies anomalies by learning substructures in attribute networks.
    \item \textbf{SL-GAD~\cite{zheng2021generative}} uses contextual views with generative attribute regression and multi-view comparative learning.
    \item \textbf{PREM~\cite{pan2023prem}} eliminates training-phase message passing through preprocessing and isochronous bit matching.
    \item \textbf{GCCAD~\cite{chen2022gccad}} proposes a pre-training strategy that generates pseudo anomalies via graph corruption to face label scarcity.
    \item \textbf{GRADATE~\cite{duan2023graph}} is a multi-view, multi-scale contrastive learning model that detects complex structural anomalies.
    \item \textbf{VGOD~\cite{huang2023unsupervised}} is a variance-based model that combines variance modeling and attribute reconstruction.
\end{itemize}
\textbf{GAE-based Methods}
\begin{itemize}[leftmargin=*]
    \item \textbf{GCNAE~\cite{kipf2016variational}} is an unsupervised model using a VGAE with GCN encoding and latent variables to embed graphs.
    \item \textbf{DOMINANT~\cite{ding2019deep}} models graph topology and attributes using GCN and reconstructs the original graph with AE.
    \item \textbf{AnomalyDAE~\cite{fan2020anomalydae}} uses dual self-encoders to capture interactions between structure and node attributes.
    \item \textbf{AdONE~\cite{bandyopadhyay2020outlier}} is an unsupervised AE approach with adversarial learning to separately learn structure and attribute.
    \item \textbf{GAD-NR~\cite{roy2024gad}} combines neighborhood reconstruction to reconstruct a node's entire neighborhood.
    \item \textbf{ADA-GAD~\cite{he2024ada}} is a two-stage anomaly denoising autoencoder that reduces anomalies, trains a graph autoencoder, and applies node regularization to prevent overfitting.
    \item \textbf{GADAM~\cite{chen2024boosting}} is a local inconsistency mining approach that resolves conflicts with GNN messaging.
\end{itemize}
\textbf{MV Methods:}
\begin{itemize}[leftmargin=*]
    \item \textbf{AnomMAN~\cite{chen2023anomman}} focuses UGAD on multi-view attribute networks using attention to fuse information across views.
    \item \textbf{DualGAD~\cite{tang2024dualgad}} combines generative and contrastive modules with masked subgraph reconstruction for UGAD.
\end{itemize}
\begin{table*}[t]
\centering
\caption{Performance comparison of all models on four small-scale datasets in the real unsupervised scenario.}
\setlength{\tabcolsep}{1.4mm}{}	
\begin{tabular}{cc|cc|cc|cc|cc}
\toprule
\multicolumn{2}{c|}{\multirow{2}{*}{Method}}  & \multicolumn{2}{c|}{Retail} & \multicolumn{2}{c|}{Alibaba}  & \multicolumn{2}{c|}{Amazon}  & \multicolumn{2}{c}{YelpChi} \\
 &  & AUC & Macro-F1 & AUC & Macro-F1 & AUC & Macro-F1 & AUC & Macro-F1 \\
\midrule
Trad. & Radar [IJCAI'17] & 0.625{\scriptsize $\pm$0.002} & 0.533{\scriptsize $\pm$0.007} & 0.659{\scriptsize $\pm$0.003} & 0.560{\scriptsize $\pm$0.008} & 0.582{\scriptsize $\pm$0.007} & 0.485{\scriptsize $\pm$0.004} & 0.502{\scriptsize $\pm$0.003} & 0.383{\scriptsize $\pm$0.011} \\
\midrule
\multirow{3}{*}{MPI} & ComGA [WSDM'22] & 0.626{\scriptsize $\pm$0.015} & 0.553{\scriptsize $\pm$0.018} & 0.591{\scriptsize $\pm$0.003} & 0.473{\scriptsize $\pm$0.006} & 0.661{\scriptsize $\pm$0.006} & 0.558{\scriptsize $\pm$0.013} & 0.535{\scriptsize $\pm$0.001} & 0.391{\scriptsize $\pm$0.001} \\
& RAND [ICDM'23] & 0.665{\scriptsize $\pm$0.007} & 0.622{\scriptsize $\pm$0.010} & 0.692{\scriptsize $\pm$0.002} & 0.603{\scriptsize $\pm$0.007} & 0.660{\scriptsize $\pm$0.004} & 0.550{\scriptsize $\pm$0.008} & 0.553{\scriptsize $\pm$0.005} & 0.401{\scriptsize $\pm$0.002} \\
& TAM [NeurIPS'24] & 0.676{\scriptsize $\pm$0.006} & 0.630{\scriptsize $\pm$0.007} & 0.712{\scriptsize $\pm$0.004} & 0.650{\scriptsize $\pm$0.008} & \underline{0.763{\scriptsize $\pm$0.011}} & 0.643{\scriptsize $\pm$0.008} & 0.570{\scriptsize $\pm$0.007} & 0.405{\scriptsize $\pm$0.008} \\
\midrule
\multirow{9}{*}{CL} & CoLA [TNNLS'21] & 0.554{\scriptsize $\pm$0.013} & 0.513{\scriptsize $\pm$0.011} & 0.570{\scriptsize $\pm$0.011} & 0.441{\scriptsize $\pm$0.018} & 0.582{\scriptsize $\pm$0.008} & 0.488{\scriptsize $\pm$0.005} & 0.764{\scriptsize $\pm$0.003} & 0.503{\scriptsize $\pm$0.007} \\
& ANEMONE [CIKM'21] & 0.630{\scriptsize $\pm$0.004} & 0.589{\scriptsize $\pm$0.009} & 0.651{\scriptsize $\pm$0.002} & 0.581{\scriptsize $\pm$0.008} & 0.622{\scriptsize $\pm$0.013} & 0.510{\scriptsize $\pm$0.014} & 0.773{\scriptsize $\pm$0.011} & 0.502{\scriptsize $\pm$0.018} \\
& Sub-CR [IJCAI'22] & 0.650{\scriptsize $\pm$0.006} & 0.594{\scriptsize $\pm$0.008} & 0.652{\scriptsize $\pm$0.017} & 0.599{\scriptsize $\pm$0.012} & 0.599{\scriptsize $\pm$0.005} & 0.493{\scriptsize $\pm$0.007} & 0.781{\scriptsize $\pm$0.012} & 0.513{\scriptsize $\pm$0.013} \\
& ARISE [TNNLS'23] & 0.671{\scriptsize $\pm$0.011} & 0.623{\scriptsize $\pm$0.017} & 0.708{\scriptsize $\pm$0.019} & 0.619{\scriptsize $\pm$0.011} & 0.692{\scriptsize $\pm$0.009} & 0.579{\scriptsize $\pm$0.006} & 0.779{\scriptsize $\pm$0.002} & 0.509{\scriptsize $\pm$0.004} \\
& SL-GAD [TKDE'21] & 0.658{\scriptsize $\pm$0.018} & 0.596{\scriptsize $\pm$0.016} & 0.700{\scriptsize $\pm$0.008} & 0.611{\scriptsize $\pm$0.009} & 0.653{\scriptsize $\pm$0.008} & 0.545{\scriptsize $\pm$0.003} & 0.798{\scriptsize $\pm$0.005} & 0.521{\scriptsize $\pm$0.007} \\
& PREM [ICDM'23] & 0.661{\scriptsize $\pm$0.008} & 0.609{\scriptsize $\pm$0.009} & 0.680{\scriptsize $\pm$0.016} & 0.607{\scriptsize $\pm$0.008} & 0.658{\scriptsize $\pm$0.018} & 0.547{\scriptsize $\pm$0.003} & 0.765{\scriptsize $\pm$0.008} & 0.503{\scriptsize $\pm$0.006} \\
& GCCAD [TKDE'22] & 0.665{\scriptsize $\pm$0.011} & 0.611{\scriptsize $\pm$0.006} & 0.701{\scriptsize $\pm$0.006} & 0.630{\scriptsize $\pm$0.004} & 0.662{\scriptsize $\pm$0.007} & 0.550{\scriptsize $\pm$0.012} & 0.772{\scriptsize $\pm$0.013} & 0.507{\scriptsize $\pm$0.007} \\
& GRADATE [AAAI'23] & 0.683{\scriptsize $\pm$0.004} & 0.644{\scriptsize $\pm$0.004} & 0.702{\scriptsize $\pm$0.007} & 0.642{\scriptsize $\pm$0.009} & 0.730{\scriptsize $\pm$0.002} & 0.616{\scriptsize $\pm$0.012} & 0.750{\scriptsize $\pm$0.001} & 0.501{\scriptsize $\pm$0.003} \\
& VGOD [ICDE'23] & 0.680{\scriptsize $\pm$0.002} & 0.639{\scriptsize $\pm$0.003} & 0.710{\scriptsize $\pm$0.013} & 0.644{\scriptsize $\pm$0.015} & 0.724{\scriptsize $\pm$0.001} & 0.606{\scriptsize $\pm$0.006} & 0.748{\scriptsize $\pm$0.009} & 0.502{\scriptsize $\pm$0.002} \\
\midrule
\multirow{7}{*}{GAE} & DOMINANT [arXiv'16] & 0.621{\scriptsize $\pm$0.009} & 0.577{\scriptsize $\pm$0.011} & 0.609{\scriptsize $\pm$0.008} & 0.549{\scriptsize $\pm$0.012} & 0.621{\scriptsize $\pm$0.005} & 0.531{\scriptsize $\pm$0.007} & 0.519{\scriptsize $\pm$0.005} & 0.380{\scriptsize $\pm$0.005} \\
& GCNAE [SDM'19] & 0.623{\scriptsize $\pm$0.007} & 0.579{\scriptsize $\pm$0.010} & 0.624{\scriptsize $\pm$0.007} & 0.564{\scriptsize $\pm$0.011} & 0.617{\scriptsize $\pm$0.017} & 0.534{\scriptsize $\pm$0.010} & 0.520{\scriptsize $\pm$0.010} & 0.385{\scriptsize $\pm$0.008} \\
& AnomalyDAE [ICASSP'20] & 0.618{\scriptsize $\pm$0.003} & 0.545{\scriptsize $\pm$0.005} & 0.665{\scriptsize $\pm$0.007} & 0.601{\scriptsize $\pm$0.008} & 0.634{\scriptsize $\pm$0.015} & 0.538{\scriptsize $\pm$0.012} & 0.573{\scriptsize $\pm$0.016} & 0.400{\scriptsize $\pm$0.010} \\
& AdONE [WSDM'20] & 0.621{\scriptsize $\pm$0.012} & 0.565{\scriptsize $\pm$0.010} & 0.641{\scriptsize $\pm$0.004} & 0.579{\scriptsize $\pm$0.004} & 0.649{\scriptsize $\pm$0.008} & 0.561{\scriptsize $\pm$0.009} & 0.670{\scriptsize $\pm$0.005} & 0.497{\scriptsize $\pm$0.003} \\
& GAD-NR [WSDM'24] & 0.681{\scriptsize $\pm$0.006} & 0.632{\scriptsize $\pm$0.005} & 0.707{\scriptsize $\pm$0.007} & 0.638{\scriptsize $\pm$0.010} & 0.754{\scriptsize $\pm$0.006} & 0.636{\scriptsize $\pm$0.005} & 0.790{\scriptsize $\pm$0.002} & 0.519{\scriptsize $\pm$0.008} \\
& ADA-GAD [AAAI'24] & 0.682{\scriptsize $\pm$0.003} & 0.645{\scriptsize $\pm$0.002} & \underline{0.715{\scriptsize $\pm$0.002}} & \underline{0.663{\scriptsize $\pm$0.002}} & 0.754{\scriptsize $\pm$0.001} & 0.639{\scriptsize $\pm$0.003} & 0.799{\scriptsize $\pm$0.006} & 0.535{\scriptsize $\pm$0.007} \\
& GADAM [ICLR'24] & \underline{0.690{\scriptsize $\pm$0.008}} & 0.644{\scriptsize $\pm$0.008} & 0.705{\scriptsize $\pm$0.001} & 0.636{\scriptsize $\pm$0.002} & 0.752{\scriptsize $\pm$0.004} & 0.633{\scriptsize $\pm$0.008} & 0.796{\scriptsize $\pm$0.005} & 0.533{\scriptsize $\pm$0.009} \\
\midrule
\multirow{2}{*}{MV} & AnomMAN [IS'23] & 0.688{\scriptsize $\pm$0.005} & 0.647{\scriptsize $\pm$0.008} & 0.710{\scriptsize $\pm$0.006} & 0.650{\scriptsize $\pm$0.009} & 0.755{\scriptsize $\pm$0.005} & \underline{0.645{\scriptsize $\pm$0.007}} & 0.801{\scriptsize $\pm$0.006} & 0.530{\scriptsize $\pm$0.004} \\
& DualGAD [IS'24] & 0.689{\scriptsize $\pm$0.007} & \underline{0.650{\scriptsize $\pm$0.003}} & 0.712{\scriptsize $\pm$0.005} & 0.653{\scriptsize $\pm$0.004} & 0.758{\scriptsize $\pm$0.002} & 0.643{\scriptsize $\pm$0.006} & \underline{0.814{\scriptsize $\pm$0.003}} & \underline{0.545{\scriptsize $\pm$0.008}} \\
\midrule
\multicolumn{2}{c|}{\textbf{UMGAD}} & \textbf{0.770{\scriptsize $\pm$0.009}} & \textbf{0.722{\scriptsize $\pm$0.005}} & \textbf{0.825{\scriptsize $\pm$0.006}} & \textbf{0.740{\scriptsize $\pm$0.006}} & \textbf{0.878{\scriptsize $\pm$0.005}} & \textbf{0.729{\scriptsize $\pm$0.004}} & \textbf{0.908{\scriptsize $\pm$0.001}} & \textbf{0.605{\scriptsize $\pm$0.004}} \\
\midrule
\multicolumn{2}{c|}{\textit{Improvement}} & $11.92\%\uparrow$ & $11.08\%\uparrow$ & $15.38\%\uparrow$ & $11.61\%\uparrow$ & $15.07\%\uparrow$ & $13.06\%\uparrow$ & $11.56\%\uparrow$ & $10.97\%\uparrow$ \\
\bottomrule
\end{tabular}
\label{tab:baselines_threshold}
\vspace{-2mm}
\end{table*}

\subsubsection{Implementation Details} We implement all baselines according to their provided code or by utilizing the PyGOD toolkit. 
Our method adopts GAT and simplified GCN as the encoder and decoder. For Amazon and YelpChi with real anomalies, the depth of the encoder and decoder is set to 2 and 1. For Retail and Alibaba with injected anomalies, the number of encoder and decoder layers is set to 1. 
AUC (area under the ROC curve) and Macro-F1 are used as performance metrics. All experiments are performed on the RTX 4090 GPU with 24GB of memory.

\subsection{Anomaly Score Threshold Selection for Unsupervised Anomaly Detection \textbf{(RQ1)}}

The results shown in Fig.~\ref{fig:threshold} illustrate the trend of ranked node anomaly scores for our \model and the best-performing baselines: ADA-GAD, TAM, GADAM, and AnomMAN on four small-scale datasets, and ADA-GAD, GRADATE, GADAM, and DualGAD on two large-scale datasets. Compared to other baselines, the curve for our \model converges more quickly, stabilizing closer to the actual number of anomalies in the datasets.
\model effectively distinguishes anomaly scores between anomalous and normal nodes, enabling more accurate anomaly identification. A robust anomaly detection model should separate anomalous nodes from normal ones using anomaly scores. When these scores are ranked in descending order, a clear inflection point should appear, where the scores stabilize. Nodes with higher scores are likely anomalous, while those with lower, more uniform scores are normal. We use the anomaly score at this inflection point as the threshold, and the number of anomalous nodes identified at this threshold closely matches the actual number of anomalies, as confirmed by ground truth. 

Our proposed anomaly threshold selection strategy is the first truly unsupervised method. It does not require any ground-truth information from the test set, making it a fair and objective model evaluation approach. Therefore, the experimental results in Table~\ref{tab:baselines_threshold} are objective. Although the results in Table~\ref{tab:baselines} are generally better than those in Table~\ref{tab:baselines_threshold}, they are not objective, as ground-truth information from the test set (e.g., the number of anomalous nodes) is used when selecting the anomaly score threshold.

\subsection{Performance Comparison in the Real Unsupervised Scenario \textbf{(RQ2)}}
Next, we evaluate the performance of our \model and all baselines in real unsupervised scenarios. 
The experimental results are shown in Table~\ref{tab:baselines_threshold}. The best results are highlighted in bold, and the second-best results are underlined. As we can see, our \model obtains the optimal performance, significantly achieves 13.48\%, and 11.68\% average improvement in terms of AUC and Macro-F1 across four datasets. We can draw the following conclusions: First, according to our proposed threshold selection strategy and the node anomaly score curves in Fig.~\ref{fig:threshold}, the number of detected anomalies corresponding to our selected thresholds is the closest to the real situation compared to other methods. Second, UMGAD considers the multi-relational correlation between nodes, and collaboratively combines GAEs with various masking mechanisms across different types of relations, effectively capturing diverse anomaly signals in both original and augmented views. Therefore, the anomaly scores of anomalous nodes can be better distinguished from those of normal nodes.

\subsection{Performance Comparison in large-sacle graphs \textbf{(RQ3)}}\label{sec.largescale}
\begin{table}[t]
\centering
\caption{Performance comparison of models on large-scale graph datasets in the real unsupervised scenario.}
\setlength{\tabcolsep}{0.9mm}{}	
\begin{tabular}{cc|cc|cc}
\toprule
\multicolumn{2}{c|}{\multirow{2}{*}{Method}}  & \multicolumn{2}{c|}{DG-Fin} & \multicolumn{2}{c}{T-Social}  \\
 &  & AUC & Macro-F1 & AUC & Macro-F1  \\
\midrule
\multirow{2}{*}{MPI} & ComGA & 0.580{\scriptsize $\pm$0.007} & 0.501{\scriptsize $\pm$0.009} & 0.549{\scriptsize $\pm$0.006} & 0.489{\scriptsize $\pm$0.005} \\
& RAND & 0.584{\scriptsize $\pm$0.009} & 0.503{\scriptsize $\pm$0.011} & 0.561{\scriptsize $\pm$0.005} & 0.498{\scriptsize $\pm$0.002} \\
\midrule
\multirow{3}{*}{CL} & PREM & 0.626{\scriptsize $\pm$0.007} & 0.510{\scriptsize $\pm$0.009} & 0.590{\scriptsize $\pm$0.003} & 0.502{\scriptsize $\pm$0.005} \\
& GRADATE & 0.636{\scriptsize $\pm$0.010} & 0.516{\scriptsize $\pm$0.011} & 0.606{\scriptsize $\pm$0.009} & 0.509{\scriptsize $\pm$0.007} \\
& VGOD & 0.632{\scriptsize $\pm$0.008} & 0.511{\scriptsize $\pm$0.010} & 0.597{\scriptsize $\pm$0.012} & 0.503{\scriptsize $\pm$0.011} \\
\midrule
\multirow{2}{*}{GAE} & ADA-GAD & 0.662{\scriptsize $\pm$0.004} & 0.530{\scriptsize $\pm$0.003} & \underline{0.653{\scriptsize $\pm$0.004}} & \underline{0.515{\scriptsize $\pm$0.004}} \\
& GADAM & 0.651{\scriptsize $\pm$0.009} & 0.522{\scriptsize $\pm$0.011} & 0.638{\scriptsize $\pm$0.011} & 0.600{\scriptsize $\pm$0.010} \\
\midrule
MV &  DualGAD & \underline{0.663{\scriptsize $\pm$0.005}} & \underline{0.532{\scriptsize $\pm$0.004}} & 0.648{\scriptsize $\pm$0.003} & 0.510{\scriptsize $\pm$0.004} \\
\midrule
\multicolumn{2}{c|}{\textbf{UMGAD}} & \textbf{0.733{\scriptsize $\pm$0.009}} & \textbf{0.598{\scriptsize $\pm$0.005}} & \textbf{0.712{\scriptsize $\pm$0.003}} & \textbf{0.558{\scriptsize $\pm$0.002}} \\
\midrule
\multicolumn{2}{c|}{\textit{Improvement}} & $10.52\%\uparrow$ & $12.52\%\uparrow$ & $9.04\%\uparrow$ & $8.49\%\uparrow$ \\
\bottomrule
\end{tabular}
\label{tab:baselines_largescale}
\end{table}

To evaluate \model on large-scale datasets DG-Fin and T-Social, we compare it with state-of-the-art methods that avoid out-of-memory (OOM) issues. Only ComGA, RAND, PREM, GRADATE, VGOD, ADA-GAD, GADAM, and DualGAD successfully run. As shown in Table~\ref{tab:baselines_largescale}, \model outperforms all models in AUC and Macro-F1, with average improvements of 9.78\% and 10.51\%, respectively, over the best baseline. DualGAD performs second, followed by ADA-GAD and GADAM. Specifically, in DGraph-Fin, only 32.2\% of users have engaged in borrowing behavior, with true anomalies making up just 1.3\%, and anomalies overall accounting for only 0.4\%. This extreme imbalance makes detection challenging. Despite this, our method consistently outperforms other SOTA methods in both effectiveness and efficiency, as shown in Table~\ref{tab:baselines_largescale} and Fig.~\ref{fig:time}. These findings highlight UMGAD's exceptional efficiency in handling large-scale graphs.

\begin{table}[h]
\centering
\caption{The comparison of UMGAD and its variants on AUC and macro-F1 (F1 for short) metrics. }
\label{tab:ablation}
\setlength{\tabcolsep}{1.4mm}{}	
\begin{tabular}{c|cc|cc|cc|cc} 
\toprule

 Dataset & \multicolumn{2}{c|}{Retail} & \multicolumn{2}{c|}{Alibaba} & \multicolumn{2}{c|}{Amazon} & \multicolumn{2}{c}{YelpChi} \\

 Metrics & AUC & F1 & AUC & F1 & AUC & F1 & AUC & F1 \\
\midrule
\midrule
\textbf{\textit{w/o M}} & 0.712 & 0.669 & 0.783 & 0.689 & 0.829 & 0.666 & 0.865 & 0.556 \\
\textbf{\textit{w/o O}} & 0.720 & 0.682 & 0.788 & 0.696 & 0.835 & 0.678 & 0.874 & 0.565 \\
\textbf{\textit{w/o A}} & 0.714 & 0.670 & 0.782 & 0.691 & 0.827 & 0.670 & 0.869 & 0.558 \\
\textbf{\textit{w/o NA}} & 0.731 & 0.690 & 0.795 & 0.706 & 0.842 & 0.686 & 0.878 & 0.575 \\
\textbf{\textit{w/o SA}} & 0.726 & 0.685 & 0.788 & 0.701 & 0.839 & 0.682 & 0.875 & 0.567 \\
\textbf{\textit{w/o DCL}} & 0.748 & 0.702 & 0.806 & 0.718 & 0.848 & 0.705 & 0.883 & 0.577 \\
\midrule
\textbf{UMGAD} & \textbf{0.770} & \textbf{0.722} & \textbf{0.825} & \textbf{0.740} & \textbf{0.878} & \textbf{0.729} & \textbf{0.908} & \textbf{0.605} \\
\bottomrule
\end{tabular}
\end{table}
\subsection{Ablation Study \textbf{(RQ4)}}
To evaluate the effectiveness of each component in \model, we further conduct ablation studies on different variants. Specifically, we generate five variants as follows:
\begin{itemize}[leftmargin=*]
    \item \textbf{\textit{w/o M}} removes the GMAE module and uses corresponding the GAE module instead.
    \item \textbf{\textit{w/o O}} removes the original-view graph-masked autoencoder and only detects augmented graphs.
    \item \textbf{\textit{w/o A}} removes the augmented-view graph-masked autoencoder and only detect original graphs.
    \item \textbf{\textit{w/o NA}} excludes the node attribute-level augmentation.
    \item \textbf{\textit{w/o SA}} excludes the subgraph-level augmentation. 
    \item \textbf{\textit{w/o DCL}} excludes the dual-view contrastive learning.
\end{itemize}

Table~\ref{tab:ablation} presents the results of model variants on four datasets in real unsupervised scenarios. All five variants underperform \model in terms of AUC and Macro-F1, underscoring the necessity and effectiveness of the proposed components. Among them, \textbf{\textit{w/o M}} performs the worst, highlighting the importance of the graph-masked autoencoder (GMAE) in GAD tasks. Ordinary GAE struggles to accurately reconstruct the attributes and structural features of anomalous nodes, leading to suboptimal anomaly detection performance. \textbf{\textit{w/o O}} outperforms \textbf{\textit{w/o M}} and \textbf{\textit{w/o A}}, emphasizing the critical role of the original multiplex heterogeneous graph. Relying solely on the augmented graph risks losing crucial anomaly information. The results for \textbf{\textit{w/o NA}} and \textbf{\textit{w/o SA}} further demonstrate that the augmented-view GMAE is effective in capturing anomalous information. Changes to node attributes and structural features in the augmented graph help mitigate the homogeneity trap common in GNN-based autoencoders. Finally, while \textbf{\textit{w/o CDL}} performs the best among all variants, it still falls short of \model, confirming the significant contribution of dual-view contrastive learning in optimizing node representations.

\begin{figure*}[t]
    \centering
    \begin{subfigure}{0.25\textwidth}
        \includegraphics[width=\linewidth]{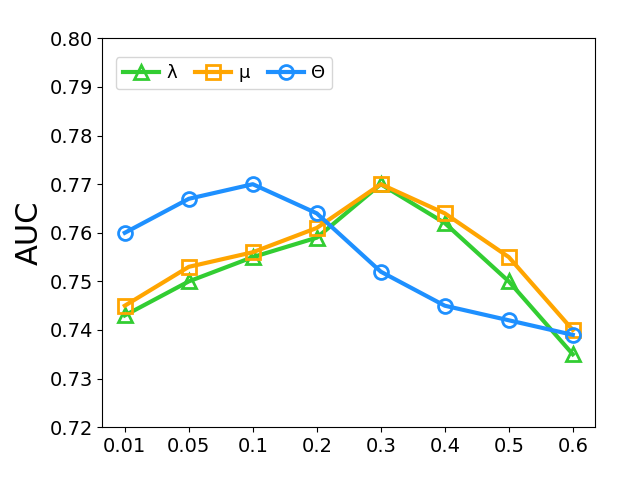}
        \vspace{-6mm}
        \caption{Retail}
        \label{fig:lambda_mu_theta_Retail}
    \end{subfigure}
    \hspace{-3mm}
    \begin{subfigure}{0.25\textwidth}
        \includegraphics[width=\linewidth]{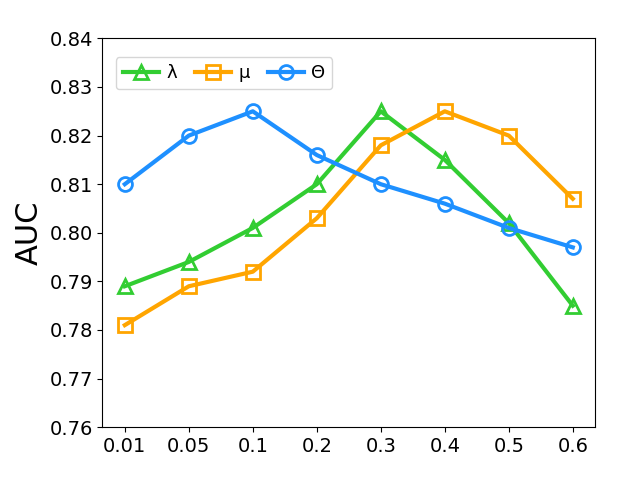}
        \vspace{-6mm}
        \caption{Alibaba}
        \label{fig:lambda_mu_theta_Alibaba}
    \end{subfigure} 
    \hspace{-3mm}
    \begin{subfigure}{0.25\textwidth}
        \includegraphics[width=\linewidth]{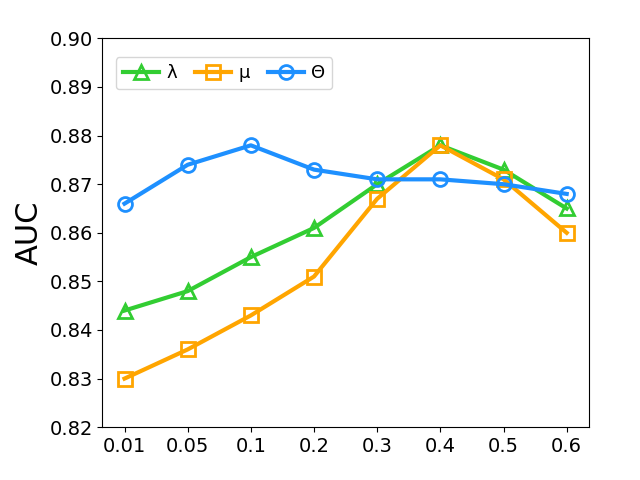}
        \vspace{-6mm}
        \caption{Amazon}
        \label{fig:lambda_mu_theta_Amazon}
    \end{subfigure}
    \hspace{-0.3cm}
    \begin{subfigure}{0.25\textwidth}
        \includegraphics[width=\linewidth]{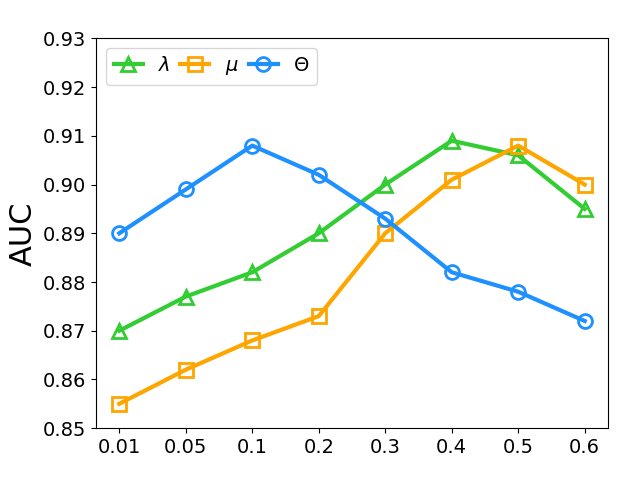}
        \vspace{-6mm}
        \caption{YelpChi}
        \label{fig:lambda_mu_theta_YelpChi}
    \end{subfigure}
    \vspace{-2mm}
    \caption{The effect of hyperparameters $\lambda$ and $\mu$ in the final loss function.}
    \label{fig:lambda_mu}
    \vspace{-2mm}
\end{figure*}
\begin{figure*}[t]
    \centering
    \begin{subfigure}{0.25\textwidth}
        \includegraphics[width=\linewidth]{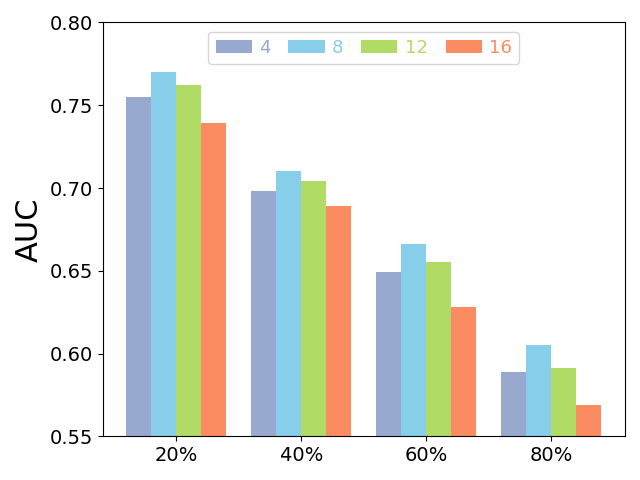}
        \vspace{-6mm}
        \caption{Retail}
        \label{fig:mask_size_Retail}
    \end{subfigure}
    \hspace{-0.3cm}
    \begin{subfigure}{0.25\textwidth}
        \includegraphics[width=\linewidth]{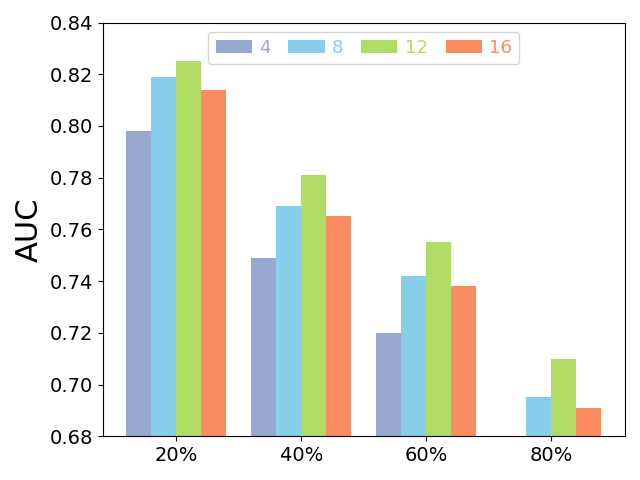}
        \vspace{-6mm}
        \caption{Alibaba}
        \label{fig:mask_size_Alibaba}
    \end{subfigure}
    \hspace{-0.3cm}
    \begin{subfigure}{0.25\textwidth}
        \includegraphics[width=\linewidth]{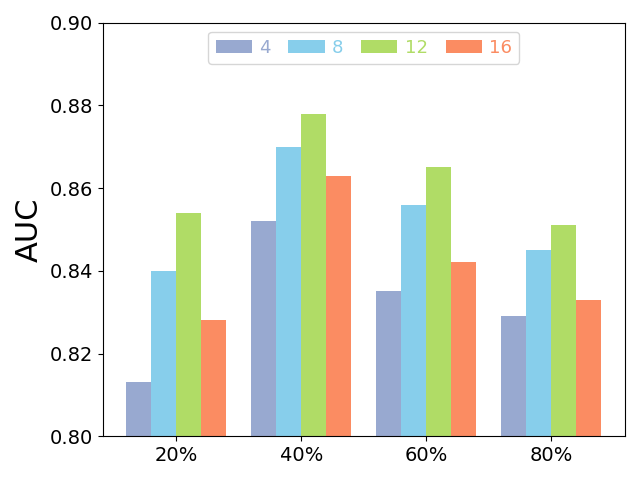}
        \vspace{-6mm}
        \caption{Amazon}
        \label{fig:mask_size_Amazon}
    \end{subfigure}
    \hspace{-0.3cm}
    \begin{subfigure}{0.25\textwidth}
        \includegraphics[width=\linewidth]{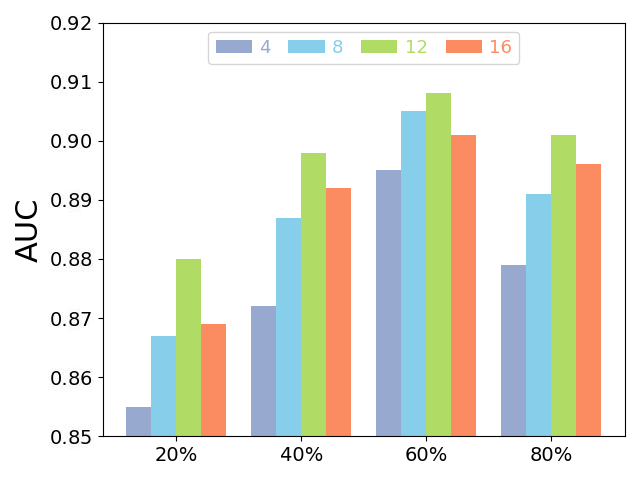}
        \vspace{-6mm}
        \caption{YelpChi}
        \label{fig:mask_size_YelpChi}
    \end{subfigure}
    \vspace{-2mm}
    \caption{Effect of masking ratio and subgraph size. X-axis represents the mask ratio, while the legend denotes subgraph size.}
    \label{fig:mask_size}
    \vspace{-2mm}
\end{figure*}

\subsection{Parameter Sensitivity Analysis \textbf{(RQ5)}}\label{sec.para}
\subsubsection{The effect of hyperparameters $\lambda$, $\mu$ and $\Theta$ in the final loss function}
$\lambda$, $\mu$ and $\Theta$ correspond to the importance parameters of the two augmented views and the dual-view contrastive learning module, respectively. Considering that incorporating node attribute-level and subgraph-level augmented views on top of the original view can improve model performance, the results of the parameter sensitivity experiments are shown in Fig.~\ref{fig:lambda_mu}. Our \model performs optimally when $\lambda$ and $\mu$ are set to 0.3 and 0.3, 0.3 and 0.4, 0.4 and 0.4, and 0.4 and 0.5 on the four small-scale datasets, respectively. Additionally, our \model consistently performs well with $\Theta$ set to 0.1 across all datasets.

\subsubsection{The effect of masking ratio $r_m$ and masking subgraph size $|\mathcal{V}_m|$}
To investigate how subgraph size $|\mathcal{V}_m|$ (number of nodes) and masking ratio $r_m$ affect model performance, we test $|\mathcal{V}_m| \in \{4, 8, 12, 16\}$ and $r_m \in \{20\%, 40\%, 60\%, 80\%\}$ on four datasets. As shown in Fig.~\ref{fig:mask_size}, YelpChi and Amazon achieve optimal performance at masking rates of 60\% and 40\%, respectively, while Retail and Alibaba perform best with a 20\% masking ratio. This difference may stem from the higher anomaly rates in YelpChi and Amazon, providing richer self-supervised signals for reconstructing masked node information. We can conclude that removing some unnecessary connections helps to anomalously reduce the amount of redundant information propagated from the normal majority.

\begin{figure}[h]
    \centering
    \vspace{-2mm}
    \begin{subfigure}{0.24\textwidth}
        \includegraphics[width=\linewidth]{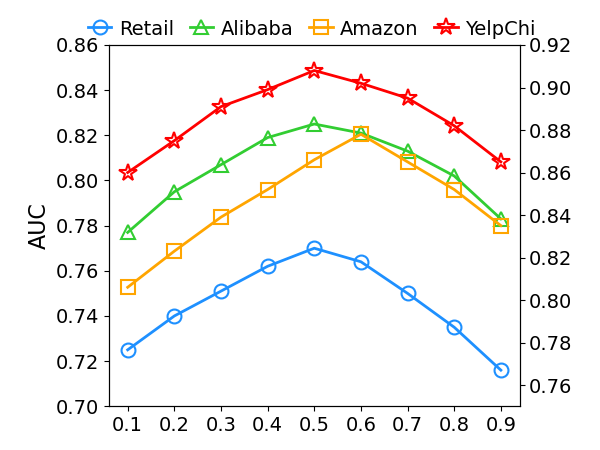}
        \vspace{-6mm}
        \caption{$\alpha$}
        \label{fig:alpha}
    \end{subfigure}
    \hspace{-2mm}
    \begin{subfigure}{0.24\textwidth}
        \includegraphics[width=\linewidth]{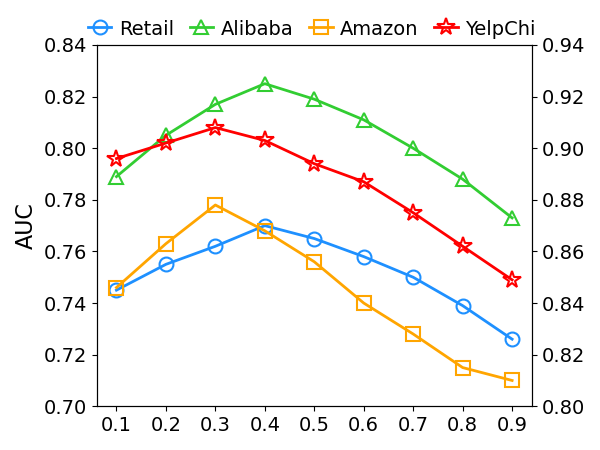}
        \vspace{-6mm}
        \caption{$\beta$}
        \label{fig:beta}
    \end{subfigure}
    \vspace{-2mm}
    \caption{The effect of weights $\alpha$ and $\beta$ in four datasets. AUC values for Retail and Alibaba are shown on the left Y-axis, while those for Amazon and YelpChi are on the right Y-axis.}
    \label{fig:alpha_beta}
\end{figure}

\subsubsection{The effect of weights $\alpha$ and $\beta$ that balance the importance of attribute and structure reconstruction}
We conduct a sensitivity analysis of $\alpha$ and $\beta$, which balance attribute and structure reconstructions in the original and augmented views. As shown in Fig.~\ref{fig:alpha_beta}, the model's performance declines sharply for extreme values of $\alpha$ and $\beta$ $(<0.2 \text{ or } >0.8)$, highlighting the importance of maintaining a balance. The AUC value first increases and then decreases as $\alpha$ and $\beta$ grow, with optimal performance of our proposed UMGAD achieved when $\alpha$ and $\beta$ are generally selected from \{0.4, 0.5, 0.6\} and \{0.3, 0.4, 0.5\} respectively, across four different datasets.

\subsubsection{The trade-off between efficiency and accuracy}
\begin{figure}[h]
    \centering
    \vspace{-2mm}
    \begin{subfigure}{0.24\textwidth}
        \includegraphics[width=\linewidth]{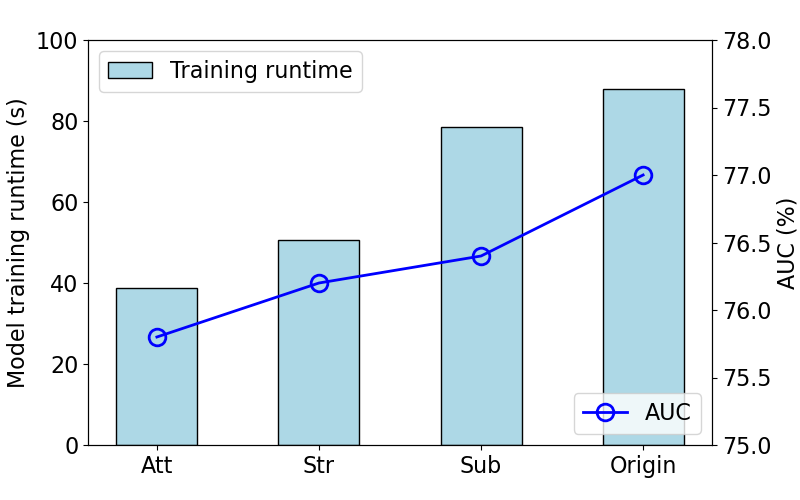}
        \vspace{-6mm}
        \caption{Retail}
        \label{fig:tradeoff_Retail}
    \end{subfigure}
    \hspace{-2mm}
    \begin{subfigure}{0.24\textwidth}
        \includegraphics[width=\linewidth]{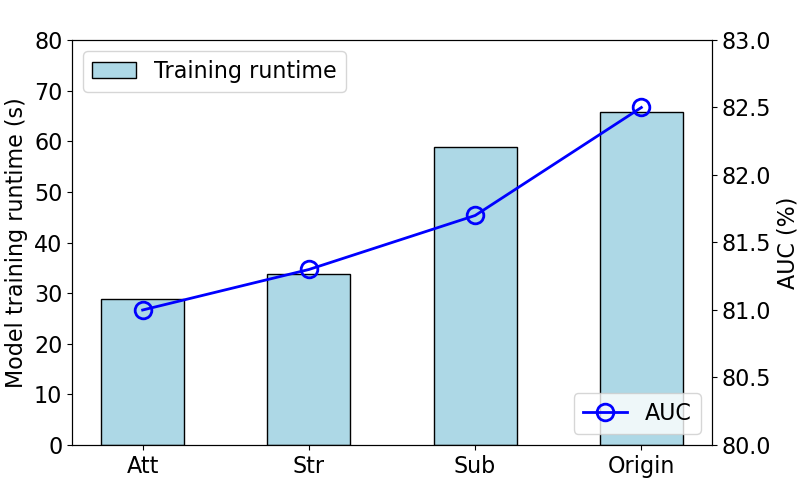}
        \vspace{-6mm}
        \caption{Alibaba}
        \label{fig:tradeoff_Alibaba}
    \end{subfigure}
    \vspace{-2mm}
    \caption{The trade-off between accuracy and efficiency on injected anomaly datasets Retail and Alibaba.}
    \label{fig:tradeoff}
    \vspace{-2mm}
\end{figure}

We evaluate the trade-off between efficiency and accuracy for our \model by analyzing its runtime and AUC scores across different variants. Specifically, \textit{\textbf{Att}} excludes structural reconstruction, retaining only attribute reconstruction in both the original and augmented views of graphs. \textit{\textbf{Str}} removes attribute reconstruction, focusing solely on structural reconstruction. \textit{\textbf{Sub}} only includes the subgraph reconstruction mechanism in both views. For \textit{\textbf{Att}} and \textit{\textbf{Str}}, we inject attribute or structural anomalies, respectively, to simulate scenarios where the anomalies are either attribute- or structure-based. Experimental results, shown in Fig.~\ref{fig:tradeoff}, reveal that by selectively pruning the model for different anomaly types, we can enhance efficiency without sacrificing significant performance, adapting to various anomaly scenarios.

\begin{table*}[t]
\centering
\caption{Performance comparison of all models on four datasets in the scenario of anomaly score threshold selection with ground truth leakage.}
\setlength{\tabcolsep}{1.4mm}{}	
\begin{tabular}{cc|cc|cc|cc|cc}
\toprule
\multicolumn{2}{c|}{\multirow{2}{*}{Method}}  & \multicolumn{2}{c|}{Retail} & \multicolumn{2}{c|}{Alibaba}  & \multicolumn{2}{c|}{Amazon}  & \multicolumn{2}{c}{YelpChi} \\
 &  & AUC & macro-F1 & AUC & macro-F1 & AUC & macro-F1 & AUC & macro-F1 \\
\midrule
Trad. & Radar [IJCAI'17] & 0.678{\scriptsize $\pm$0.003} & 0.570{\scriptsize $\pm$0.001} & 0.695{\scriptsize $\pm$0.011} & 0.621{\scriptsize $\pm$0.009} & 0.659{\scriptsize $\pm$0.003} & 0.543{\scriptsize $\pm$0.002} & 0.515{\scriptsize $\pm$0.003} & 0.402{\scriptsize $\pm$0.005} \\
\midrule
\multirow{3}{*}{MPI} & ComGA [WSDM'22] & 0.702{\scriptsize $\pm$0.009} & 0.648{\scriptsize $\pm$0.003} & 0.661{\scriptsize $\pm$0.008} & 0.552{\scriptsize $\pm$0.002} & 0.715{\scriptsize $\pm$0.004} & 0.602{\scriptsize $\pm$0.007} & 0.543{\scriptsize $\pm$0.011} & 0.397{\scriptsize $\pm$0.006} \\
& RAND [ICDM'23] & 0.726{\scriptsize $\pm$0.006} & 0.678{\scriptsize $\pm$0.007} & 0.784{\scriptsize $\pm$0.008} & 0.677{\scriptsize $\pm$0.004} & 0.730{\scriptsize $\pm$0.013} & 0.615{\scriptsize $\pm$0.006} & 0.554{\scriptsize $\pm$0.007} & 0.404{\scriptsize $\pm$0.008} \\
& TAM [NeurIPS'24] & \underline{0.753{\scriptsize $\pm$0.001}} & 0.703{\scriptsize $\pm$0.001} & 0.790{\scriptsize $\pm$0.001} & 0.715{\scriptsize $\pm$0.001} & \underline{0.848{\scriptsize $\pm$0.004}} & 0.703{\scriptsize $\pm$0.001} & 0.582{\scriptsize $\pm$0.001} & 0.425{\scriptsize $\pm$0.001} \\
\midrule
\multirow{9}{*}{CL} & CoLA [TNNLS'21] & 0.629{\scriptsize $\pm$0.009} & 0.559{\scriptsize $\pm$0.006} & 0.647{\scriptsize $\pm$0.014} & 0.553{\scriptsize $\pm$0.003} & 0.661{\scriptsize $\pm$0.006} & 0.554{\scriptsize $\pm$0.002} & 0.792{\scriptsize $\pm$0.010} & 0.519{\scriptsize $\pm$0.009} \\
& ANEMONE [CIKM'21] & 0.689{\scriptsize $\pm$0.004} & 0.631{\scriptsize $\pm$0.004} & 0.710{\scriptsize $\pm$0.007} & 0.632{\scriptsize $\pm$0.009} & 0.704{\scriptsize $\pm$0.002} & 0.584{\scriptsize $\pm$0.005} & 0.807{\scriptsize $\pm$0.006} & 0.531{\scriptsize $\pm$0.003} \\
& Sub-CR [IJCAI'22] & 0.704{\scriptsize $\pm$0.009} & 0.635{\scriptsize $\pm$0.008} & 0.728{\scriptsize $\pm$0.006} & 0.648{\scriptsize $\pm$0.008} & 0.675{\scriptsize $\pm$0.018} & 0.568{\scriptsize $\pm$0.003} & 0.810{\scriptsize $\pm$0.009} & 0.539{\scriptsize $\pm$0.008} \\
& ARISE [TNNLS'23] & 0.740{\scriptsize $\pm$0.018} & 0.683{\scriptsize $\pm$0.005} & 0.774{\scriptsize $\pm$0.006} & 0.669{\scriptsize $\pm$0.003} & 0.768{\scriptsize $\pm$0.006} & 0.645{\scriptsize $\pm$0.002} & 0.815{\scriptsize $\pm$0.001} & 0.535{\scriptsize $\pm$0.007} \\
& SL-GAD [TKDE'21] & 0.725{\scriptsize $\pm$0.003} & 0.667{\scriptsize $\pm$0.007} & 0.778{\scriptsize $\pm$0.005} & 0.672{\scriptsize $\pm$0.003} & 0.725{\scriptsize $\pm$0.009} & 0.610{\scriptsize $\pm$0.010} & 0.825{\scriptsize $\pm$0.013} & 0.540{\scriptsize $\pm$0.005} \\
& PREM [ICDM'23] & 0.724{\scriptsize $\pm$0.001} & 0.675{\scriptsize $\pm$0.003} & 0.782{\scriptsize $\pm$0.011} & 0.670{\scriptsize $\pm$0.009} & 0.735{\scriptsize $\pm$0.007} & 0.619{\scriptsize $\pm$0.006} & 0.787{\scriptsize $\pm$0.003} & 0.513{\scriptsize $\pm$0.007} \\
& GCCAD [TKDE'22] & 0.731{\scriptsize $\pm$0.003} & 0.677{\scriptsize $\pm$0.005} & 0.788{\scriptsize $\pm$0.002} & 0.689{\scriptsize $\pm$0.003} & 0.733{\scriptsize $\pm$0.001} & 0.617{\scriptsize $\pm$0.005} & 0.796{\scriptsize $\pm$0.015} & 0.516{\scriptsize $\pm$0.006} \\
& GRADATE [AAAI'23] & 0.746{\scriptsize $\pm$0.008} & 0.686{\scriptsize $\pm$0.005} & 0.790{\scriptsize $\pm$0.014} & 0.705{\scriptsize $\pm$0.013} & 0.805{\scriptsize $\pm$0.007} & 0.685{\scriptsize $\pm$0.005} & 0.780{\scriptsize $\pm$0.005} & 0.514{\scriptsize $\pm$0.006} \\
& VGOD [ICDE'23] & 0.745{\scriptsize $\pm$0.002} & 0.679{\scriptsize $\pm$0.003} & 0.791{\scriptsize $\pm$0.005} & 0.703{\scriptsize $\pm$0.003} & 0.797{\scriptsize $\pm$0.001} & 0.677{\scriptsize $\pm$0.006} & 0.771{\scriptsize $\pm$0.006} & 0.512{\scriptsize $\pm$0.008} \\
\midrule
\multirow{7}{*}{GAE} & DOMINANT [arXiv'16] & 0.681{\scriptsize $\pm$0.013} & 0.625{\scriptsize $\pm$0.011} & 0.699{\scriptsize $\pm$0.019} & 0.615{\scriptsize $\pm$0.011} & 0.694{\scriptsize $\pm$0.007} & 0.598{\scriptsize $\pm$0.005} & 0.539{\scriptsize $\pm$0.003} & 0.389{\scriptsize $\pm$0.003} \\
& GCNAE [SDM'19] & 0.683{\scriptsize $\pm$0.005} & 0.633{\scriptsize $\pm$0.006} & 0.701{\scriptsize $\pm$0.017} & 0.631{\scriptsize $\pm$0.019} & 0.679{\scriptsize $\pm$0.010} & 0.591{\scriptsize $\pm$0.012} & 0.541{\scriptsize $\pm$0.009} & 0.401{\scriptsize $\pm$0.006} \\
& AnomalyDAE [ICASSP'20] & 0.679{\scriptsize $\pm$0.009} & 0.622{\scriptsize $\pm$0.004} & 0.697{\scriptsize $\pm$0.023} & 0.626{\scriptsize $\pm$0.018} & 0.708{\scriptsize $\pm$0.018} & 0.603{\scriptsize $\pm$0.024} & 0.583{\scriptsize $\pm$0.035} & 0.419{\scriptsize $\pm$0.007} \\
& AdONE [WSDM'20] & 0.685{\scriptsize $\pm$0.009} & 0.624{\scriptsize $\pm$0.004} & 0.707{\scriptsize $\pm$0.020} & 0.637{\scriptsize $\pm$0.018} & 0.721{\scriptsize $\pm$0.015} & 0.616{\scriptsize $\pm$0.014} & 0.686{\scriptsize $\pm$0.011} & 0.506{\scriptsize $\pm$0.018} \\
& GAD-NR [WSDM'24] & 0.749{\scriptsize $\pm$0.002} & 0.691{\scriptsize $\pm$0.003} & 0.789{\scriptsize $\pm$0.013} & 0.710{\scriptsize $\pm$0.015} & 0.829{\scriptsize $\pm$0.001} & 0.703{\scriptsize $\pm$0.006} & 0.801{\scriptsize $\pm$0.002} & 0.530{\scriptsize $\pm$0.008} \\
& ADA-GAD [AAAI'24] & 0.742{\scriptsize $\pm$0.009} & 0.702{\scriptsize $\pm$0.010} & 0.795{\scriptsize $\pm$0.003} & \underline{0.716{\scriptsize $\pm$0.004}} & 0.837{\scriptsize $\pm$0.016} & 0.706{\scriptsize $\pm$0.014} & 0.832{\scriptsize $\pm$0.003} & 0.552{\scriptsize $\pm$0.006} \\
& GADAM [ICLR'24] & 0.751{\scriptsize $\pm$0.001} & 0.702{\scriptsize $\pm$0.003} & \underline{0.796{\scriptsize $\pm$0.009}} & 0.713{\scriptsize $\pm$0.011} & 0.832{\scriptsize $\pm$0.003} & 0.704{\scriptsize $\pm$0.001} & 0.821{\scriptsize $\pm$0.005} & 0.549{\scriptsize $\pm$0.005} \\
\midrule
\multirow{2}{*}{MV} & AnomMAN [IS'23] & 0.750{\scriptsize $\pm$0.008} & \underline{0.703{\scriptsize $\pm$0.006}} & 0.792{\scriptsize $\pm$0.006} & 0.714{\scriptsize $\pm$0.007} & 0.841{\scriptsize $\pm$0.007} & \underline{0.711{\scriptsize $\pm$0.005}} & 0.837{\scriptsize $\pm$0.011} & 0.554{\scriptsize $\pm$0.010} \\
& DualGAD [IS'23] & 0.752{\scriptsize $\pm$0.005} & 0.700{\scriptsize $\pm$0.008} & 0.789{\scriptsize $\pm$0.013} & 0.714{\scriptsize $\pm$0.014} & 0.839{\scriptsize $\pm$0.008} & 0.708{\scriptsize $\pm$0.004} & \underline{0.849{\scriptsize $\pm$0.006}} & \underline{0.572{\scriptsize $\pm$0.014}} \\
\midrule
\multicolumn{2}{c|}{\textbf{UMGAD}} & \textbf{0.780{\scriptsize $\pm$0.004}} & \textbf{0.728{\scriptsize $\pm$0.007}} & \textbf{0.831{\scriptsize $\pm$0.003}} & \textbf{0.744{\scriptsize $\pm$0.006}} & \textbf{0.886{\scriptsize $\pm$0.004}} & \textbf{0.749{\scriptsize $\pm$0.002}} & \textbf{0.912{\scriptsize $\pm$0.002}} & \textbf{0.610{\scriptsize $\pm$0.004}} \\
\midrule
\multicolumn{2}{c|}{\textit{Improvement}} & $3.59\%\uparrow$ & $3.56\%\uparrow$ & $4.40\%\uparrow$ & $3.91\%\uparrow$ & $4.48\%\uparrow$ & $8.33\%\uparrow$ & $7.42\%\uparrow$ & $6.64\%\uparrow$ \\
\bottomrule
\end{tabular}
\label{tab:baselines}
\vspace{-1mm}
\end{table*}

\begin{figure*}[t]
    \centering
    \begin{subfigure}{0.31\textwidth}
        \includegraphics[width=\linewidth]{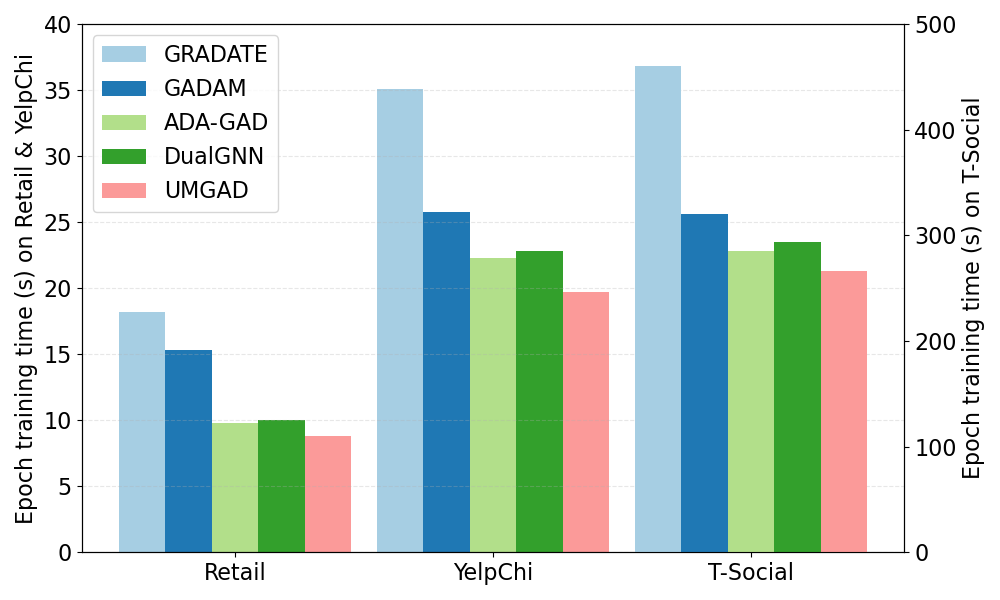}
        \vspace{-6mm}
        \caption{Running time of each epoch.}
        \label{fig:et}
    \end{subfigure}
    \hspace{-1mm}
    \begin{subfigure}{0.31\textwidth}
        \includegraphics[width=\linewidth]{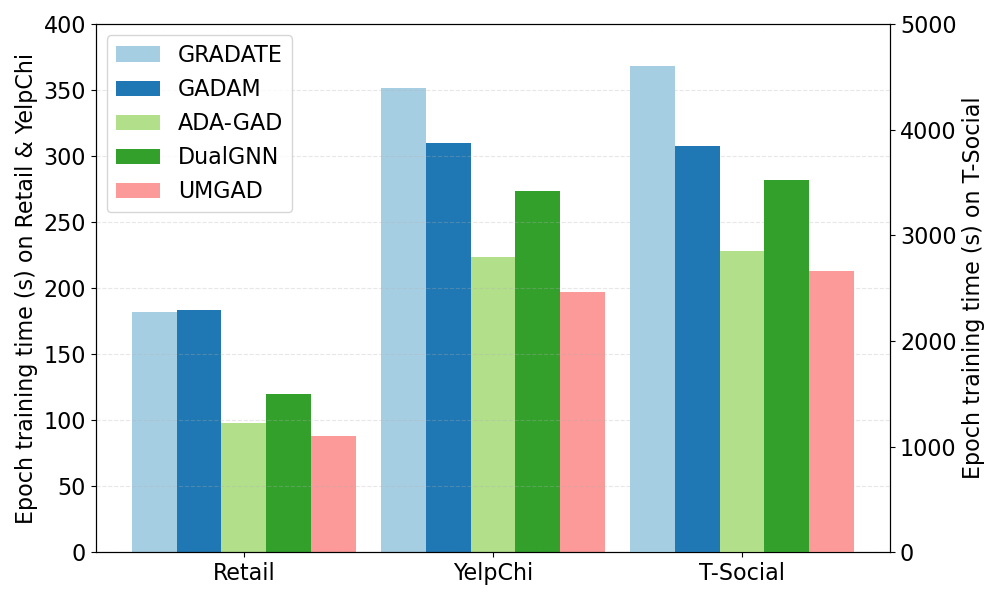}
        \vspace{-6mm}
        \caption{Running time of total model.}
        \label{fig:tt}
    \end{subfigure}
    \hspace{-1mm}
    \begin{subfigure}{0.343\textwidth}
        \includegraphics[width=\linewidth]{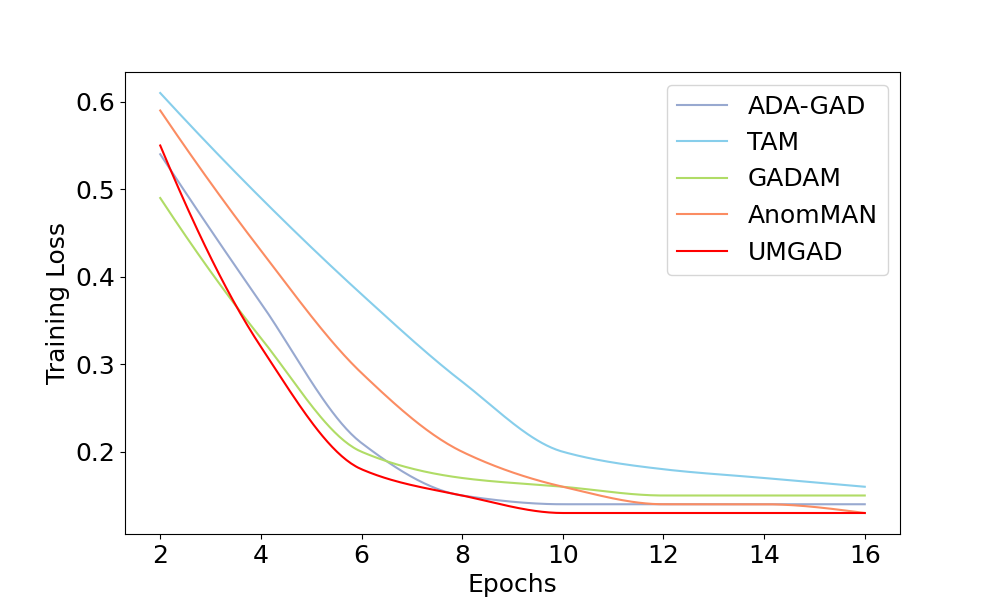}
        \vspace{-6mm}
        \caption{Training loss on YelpChi.}
        \label{fig:tl}
    \end{subfigure}
    \caption{Efficiency analysis of methods.}
    \label{fig:time}
    \vspace{-1mm}
\end{figure*}

\subsection{Performance Comparison in the Ground Truth Leakage Scenario \textbf{(RQ6)}} 

In this experiment, following~\cite{chen2023anomman}, we determine the anomaly score thresholds for all methods on each dataset based on the number of anomalies. The experimental results are presented in Table~\ref{tab:baselines}. As shown in Table~\ref{tab:baselines}, even when ground truth information is incorporated, our model still achieves the best performance. This demonstrates that \model effectively captures anomaly information and identifies anomalous nodes with greater accuracy compared to other methods.

For traditional GAD methods, Radar utilizes residual networks to learn node attributes and capture anomalies simply and effectively. However, it struggles to capture anomalous features when dealing with complex network structures. Among specifically designed MPI methods, TAM outperforms ComGA and RAND by learning custom node embeddings. It maximizes the local affinity between nodes and their neighbors, while minimizing the influence of non-homogeneous edges linking abnormal and normal nodes.

CL-based GAD methods, such as GRADATE, VGOD, and ARISE, stand out as the most effective approaches. All three methods emphasize the importance of anomalous structural subgraphs in the GAD problem, leveraging subgraph sampling strategies to learn subgraph features and capture structural anomalies that are difficult for other methods to detect. In the realm of GAE-based GAD methods, most focus on improving the detection of anomalous nodes by learning better features during the reconstruction of attribute and structural information. The four most effective and representative methods in this category are GADAM, AnomMAN, GAD-NR, and ADA-GAD. ADA-GAD enhances anomalous node feature learning through graph augmentation, GAD-NR incorporates neighborhood reconstruction into the GAD task, and GADAM improves anomaly detection performance through enhanced GNN message passing. However, these methods fall short when addressing the GAD problem on multiplex heterogeneous graphs, leading to slightly lower performance compared to our \model. AnomMAN focuses on multi-view graphs derived from ordinary heterogeneous graphs, similar to our \model. However, its performance is limited by its inability to effectively capture subgraph structural anomalies.

\subsection{Model Efficiency Analysis (\textbf{RQ7})}
To evaluate the time efficiency of \model, we compare its single-epoch and total runtime across Retail, YelpChi, and T-Social datasets against four best-performing SOTA models: GRADATE, GADAM, ADA-GAD, and DualGAN. As shown in Fig.~\ref{fig:time}, \model is faster than other baselines regarding single-epoch and total runtime across all datasets. Significantly, on the large-scale dataset T-Social, \model achieves the shortest epoch and total runtime, highlighting its superior efficiency. Additionally, as shown in Fig.~\ref{fig:tl}, \model demonstrates rapid convergence during training, reaching optimal performance with fewer epochs. This combination of efficiency and robust anomaly detection performance underscores \model's practicality and scalability, making it a probable choice for large-scale, time-sensitive graph analysis tasks.
\section{Conclusion}
In this paper, we propose \model, an unsupervised framework for anomaly detection on multiplex heterogeneous graphs. It combines different relation-aware masking mechanisms with graph autoencoders to capture informative signals while reducing redundancy. Attribute and structure reconstruction on both original and augmented views, followed by dual-view contrastive learning, ensures robust anomaly detection. Extensive experiments on datasets with injected and real-world anomalies show that \model outperforms state-of-the-art methods in accuracy and efficiency.

\section*{acknowledge}
This work is partially supported by the National Natural Science Foundation of China under grant Nos. 62176243 and 62472263, the National Science and Technology Major Project under grant No. 2024ZD0607700, the China Postdoctoral Science Foundation (No. 2024T170868), the Postdoctoral Fellowship Program of CPSF (No. GZC20232500), and the Shandong Postdoctoral Science Foundation (No. SDCX-ZG-202400317).

\clearpage
\balance

\bibliographystyle{IEEEtran}
\bibliography{ICDE}

\begin{thebibliography}{10}
\providecommand{\url}[1]{#1}
\csname url@samestyle\endcsname
\providecommand{\newblock}{\relax}
\providecommand{\bibinfo}[2]{#2}
\providecommand{\BIBentrySTDinterwordspacing}{\spaceskip=0pt\relax}
\providecommand{\BIBentryALTinterwordstretchfactor}{4}
\providecommand{\BIBentryALTinterwordspacing}{\spaceskip=\fontdimen2\font plus
\BIBentryALTinterwordstretchfactor\fontdimen3\font minus \fontdimen4\font\relax}
\providecommand{\BIBforeignlanguage}[2]{{%
\expandafter\ifx\csname l@#1\endcsname\relax
\typeout{** WARNING: IEEEtran.bst: No hyphenation pattern has been}%
\typeout{** loaded for the language `#1'. Using the pattern for}%
\typeout{** the default language instead.}%
\else
\language=\csname l@#1\endcsname
\fi
#2}}
\providecommand{\BIBdecl}{\relax}
\BIBdecl

\bibitem{ding2021inductive}
K.~Ding, J.~Li, N.~Agarwal, and H.~Liu, ``Inductive anomaly detection on attributed networks,'' in \emph{IJCAI}, 2021, pp. 1288--1294.

\bibitem{ma2021comprehensive}
X.~Ma, J.~Wu, S.~Xue, J.~Yang, C.~Zhou, Q.~Z. Sheng, H.~Xiong, and L.~Akoglu, ``A comprehensive survey on graph anomaly detection with deep learning,'' \emph{IEEE TKDE}, vol.~35, no.~12, pp. 12\,012--12\,038, 2021.

\bibitem{chai2022can}
Z.~Chai, S.~You, Y.~Yang, S.~Pu, J.~Xu, H.~Cai, and W.~Jiang, ``Can abnormality be detected by graph neural networks?'' in \emph{IJCAI}, 2022, pp. 1945--1951.

\bibitem{huang2022dgraph}
X.~Huang, Y.~Yang, Y.~Wang, C.~Wang, Z.~Zhang, J.~Xu, L.~Chen, and M.~Vazirgiannis, ``Dgraph: A large-scale financial dataset for graph anomaly detection,'' \emph{NeurIPS}, vol.~35, pp. 22\,765--22\,777, 2022.

\bibitem{dou2020enhancing}
Y.~Dou, Z.~Liu, L.~Sun, Y.~Deng, H.~Peng, and P.~S. Yu, ``Enhancing graph neural network-based fraud detectors against camouflaged fraudsters,'' in \emph{CIKM}, 2020, pp. 315--324.

\bibitem{liu2021intention}
C.~Liu, L.~Sun, X.~Ao, J.~Feng, Q.~He, and H.~Yang, ``Intention-aware heterogeneous graph attention networks for fraud transactions detection,'' in \emph{SIGKDD}, 2021, pp. 3280--3288.

\bibitem{liu2021pick}
Y.~Liu, X.~Ao, Z.~Qin, J.~Chi, J.~Feng, H.~Yang, and Q.~He, ``Pick and choose: a gnn-based imbalanced learning approach for fraud detection,'' in \emph{WWW}, 2021, pp. 3168--3177.

\bibitem{chen2022antibenford}
T.~Chen and C.~Tsourakakis, ``Antibenford subgraphs: Unsupervised anomaly detection in financial networks,'' in \emph{SIGKDD}, 2022, pp. 2762--2770.

\bibitem{yang2019mining}
Y.~Yang, Y.~Xu, Y.~Sun, Y.~Dong, F.~Wu, and Y.~Zhuang, ``Mining fraudsters and fraudulent strategies in large-scale mobile social networks,'' \emph{IEEE TKDE}, vol.~33, no.~1, pp. 169--179, 2019.

\bibitem{cheng2021causal}
L.~Cheng, R.~Guo, K.~Shu, and H.~Liu, ``Causal understanding of fake news dissemination on social media,'' in \emph{SIGKDD}, 2021, pp. 148--157.

\bibitem{min2022divide}
E.~Min, Y.~Rong, Y.~Bian, T.~Xu, P.~Zhao, J.~Huang, and S.~Ananiadou, ``Divide-and-conquer: Post-user interaction network for fake news detection on social media,'' in \emph{WWW}, 2022, pp. 1148--1158.

\bibitem{cao2024hierarchical}
Y.~Cao, H.~Peng, Z.~Yu, and S.~Y. Philip, ``Hierarchical and incremental structural entropy minimization for unsupervised social event detection,'' in \emph{AAAI}, vol.~38, no.~8, 2024, pp. 8255--8264.

\bibitem{tang2024dualgad}
H.~Tang, X.~Liang, J.~Wang, and S.~Zhang, ``Dualgad: Dual-bootstrapped self-supervised learning for graph anomaly detection,'' \emph{Information Sciences}, vol. 668, p. 120520, 2024.

\bibitem{gong2023beyond}
Z.~Gong, G.~Wang, Y.~Sun, Q.~Liu, Y.~Ning, H.~Xiong, and J.~Peng, ``Beyond homophily: Robust graph anomaly detection via neural sparsification.'' in \emph{IJCAI}, 2023, pp. 2104--2113.

\bibitem{zhang2024dig}
J.~Zhang, Z.~Xu, D.~Lv, Z.~Shi, D.~Shen, J.~Jin, and F.~Dong, ``Dig-in-gnn: Discriminative feature guided gnn-based fraud detector against inconsistencies in multi-relation fraud graph,'' in \emph{AAAI}, vol.~38, no.~8, 2024, pp. 9323--9331.

\bibitem{yang2023ahead}
S.~Yang, B.~Zhang, S.~Feng, Z.~Tan, Q.~Zheng, J.~Zhou, and M.~Luo, ``Ahead: A triple attention based heterogeneous graph anomaly detection approach,'' in \emph{Chinese Intelligent Automation Conference}.\hskip 1em plus 0.5em minus 0.4em\relax Springer, 2023, pp. 542--552.

\bibitem{gao2023addressing}
Y.~Gao, X.~Wang, X.~He, Z.~Liu, H.~Feng, and Y.~Zhang, ``Addressing heterophily in graph anomaly detection: A perspective of graph spectrum,'' in \emph{WWW}, 2023, pp. 1528--1538.

\bibitem{xu2024revisiting}
F.~Xu, N.~Wang, H.~Wu, X.~Wen, X.~Zhao, and H.~Wan, ``Revisiting graph-based fraud detection in sight of heterophily and spectrum,'' in \emph{AAAI}, vol.~38, no.~8, 2024, pp. 9214--9222.

\bibitem{zhang2022dual}
G.~Zhang, Z.~Yang, J.~Wu, J.~Yang, S.~Xue, H.~Peng, J.~Su, C.~Zhou, Q.~Z. Sheng, L.~Akoglu \emph{et~al.}, ``Dual-discriminative graph neural network for imbalanced graph-level anomaly detection,'' \emph{NeurIPS}, vol.~35, pp. 24\,144--24\,157, 2022.

\bibitem{tang2022rethinking}
J.~Tang, J.~Li, Z.~Gao, and J.~Li, ``Rethinking graph neural networks for anomaly detection,'' in \emph{ICML}.\hskip 1em plus 0.5em minus 0.4em\relax PMLR, 2022, pp. 21\,076--21\,089.

\bibitem{zhou2023improving}
S.~Zhou, X.~Huang, N.~Liu, H.~Zhou, F.-L. Chung, and L.-K. Huang, ``Improving generalizability of graph anomaly detection models via data augmentation,'' \emph{IEEE TKDE}, vol.~35, no.~12, pp. 12\,721--12\,735, 2023.

\bibitem{liu2022dagad}
F.~Liu, X.~Ma, J.~Wu, J.~Yang, S.~Xue, A.~Beheshti, C.~Zhou, H.~Peng, Q.~Z. Sheng, and C.~C. Aggarwal, ``Dagad: Data augmentation for graph anomaly detection,'' in \emph{ICDM}.\hskip 1em plus 0.5em minus 0.4em\relax IEEE, 2022, pp. 259--268.

\bibitem{xie2023unsupervised}
Z.~Xie, H.~Xu, W.~Chen, W.~Li, H.~Jiang, L.~Su, H.~Wang, and D.~Pei, ``Unsupervised anomaly detection on microservice traces through graph vae,'' in \emph{WWW}, 2023, pp. 2874--2884.

\bibitem{xu2024unsupervised}
B.~Xu, J.~Wang, Z.~Zhao, H.~Lin, and F.~Xia, ``Unsupervised anomaly detection on attributed networks with graph contrastive learning for consumer electronics security,'' \emph{IEEE TCE}, 2024.

\bibitem{zhang2023graph}
F.~Zhang, S.~Kan, D.~Zhang, Y.~Cen, L.~Zhang, and V.~Mladenovic, ``A graph model-based multiscale feature fitting method for unsupervised anomaly detection,'' \emph{Pattern Recognition}, vol. 138, p. 109373, 2023.

\bibitem{wang2023unsupervised}
N.~Wang, Z.~Wang, Y.~Gong, Z.~Huang, Z.~Huang, X.~Wen, and H.~Zeng, ``Unsupervised data anomaly detection based on graph neural network,'' in \emph{ICCSIA}.\hskip 1em plus 0.5em minus 0.4em\relax Springer, 2023, pp. 552--564.

\bibitem{li2017radar}
J.~Li, H.~Dani, X.~Hu, and H.~Liu, ``Radar: Residual analysis for anomaly detection in attributed networks.'' in \emph{IJCAI}, vol.~17, 2017, pp. 2152--2158.

\bibitem{luo2022comga}
X.~Luo, J.~Wu, A.~Beheshti, J.~Yang, X.~Zhang, Y.~Wang, and S.~Xue, ``Comga: Community-aware attributed graph anomaly detection,'' in \emph{WSDM}, 2022, pp. 657--665.

\bibitem{qiao2024truncated}
H.~Qiao and G.~Pang, ``Truncated affinity maximization: One-class homophily modeling for graph anomaly detection,'' \emph{NeurIPS}, vol.~36, 2024.

\bibitem{chen2022gccad}
B.~Chen, J.~Zhang, X.~Zhang, Y.~Dong, J.~Song, P.~Zhang, K.~Xu, E.~Kharlamov, and J.~Tang, ``Gccad: Graph contrastive coding for anomaly detection,'' \emph{IEEE TKDE}, vol.~35, no.~8, pp. 8037--8051, 2022.

\bibitem{duan2023graph}
J.~Duan, S.~Wang, P.~Zhang, E.~Zhu, J.~Hu, H.~Jin, Y.~Liu, and Z.~Dong, ``Graph anomaly detection via multi-scale contrastive learning networks with augmented view,'' in \emph{AAAI}, vol.~37, no.~6, 2023, pp. 7459--7467.

\bibitem{huang2023unsupervised}
Y.~Huang, L.~Wang, F.~Zhang, and X.~Lin, ``Unsupervised graph outlier detection: Problem revisit, new insight, and superior method,'' in \emph{ICDE}.\hskip 1em plus 0.5em minus 0.4em\relax IEEE, 2023, pp. 2565--2578.

\bibitem{he2024ada}
J.~He, Q.~Xu, Y.~Jiang, Z.~Wang, and Q.~Huang, ``Ada-gad: Anomaly-denoised autoencoders for graph anomaly detection,'' in \emph{AAAI}, vol.~38, no.~8, 2024, pp. 8481--8489.

\bibitem{liu2020fast}
Z.~Liu, C.~Huang, Y.~Yu, B.~Fan, and J.~Dong, ``Fast attributed multiplex heterogeneous network embedding,'' in \emph{CIKM}, 2020, pp. 995--1004.

\bibitem{yu2022multiplex}
P.~Yu, C.~Fu, Y.~Yu, C.~Huang, Z.~Zhao, and J.~Dong, ``Multiplex heterogeneous graph convolutional network,'' in \emph{SIGKDD}, 2022, pp. 2377--2387.

\bibitem{li2025dual}
X.~Li, C.~Fu, Z.~Zhao, G.~Zheng, C.~Huang, Y.~Yu, and J.~Dong, ``Dual-channel multiplex graph neural networks for recommendation,'' \emph{IEEE TKDE}, 2025.

\bibitem{pei2022resgcn}
Y.~Pei, T.~Huang, W.~van Ipenburg, and M.~Pechenizkiy, ``Resgcn: attention-based deep residual modeling for anomaly detection on attributed networks,'' \emph{Machine Learning}, vol. 111, no.~2, pp. 519--541, 2022.

\bibitem{wang2021modeling}
L.~Wang, P.~Li, K.~Xiong, J.~Zhao, and R.~Lin, ``Modeling heterogeneous graph network on fraud detection: A community-based framework with attention mechanism,'' in \emph{CIKM}, 2021, pp. 1959--1968.

\bibitem{bei2023reinforcement}
Y.~Bei, S.~Zhou, Q.~Tan, H.~Xu, H.~Chen, Z.~Li, and J.~Bu, ``Reinforcement neighborhood selection for unsupervised graph anomaly detection,'' in \emph{ICDM}.\hskip 1em plus 0.5em minus 0.4em\relax IEEE, 2023, pp. 11--20.

\bibitem{liu2024towards}
Y.~Liu, K.~Ding, Q.~Lu, F.~Li, L.~Y. Zhang, and S.~Pan, ``Towards self-interpretable graph-level anomaly detection,'' \emph{NeurIPS}, vol.~36, 2024.

\bibitem{kong2024federated}
X.~Kong, W.~Zhang, H.~Wang, M.~Hou, X.~Chen, X.~Yan, and S.~K. Das, ``Federated graph anomaly detection via contrastive self-supervised learning,'' \emph{IEEE TNNLS}, 2024.

\bibitem{liu2024bourne}
J.~Liu, M.~He, X.~Shang, J.~Shi, B.~Cui, and H.~Yin, ``Bourne: Bootstrapped self-supervised learning framework for unified graph anomaly detection,'' in \emph{ICDE}.\hskip 1em plus 0.5em minus 0.4em\relax IEEE, 2024, pp. 2820--2833.

\bibitem{wang2023cross}
Q.~Wang, G.~Pang, M.~Salehi, W.~Buntine, and C.~Leckie, ``Cross-domain graph anomaly detection via anomaly-aware contrastive alignment,'' in \emph{AAAI}, vol.~37, no.~4, 2023, pp. 4676--4684.

\bibitem{liu2021anomaly}
Y.~Liu, Z.~Li, S.~Pan, C.~Gong, C.~Zhou, and G.~Karypis, ``Anomaly detection on attributed networks via contrastive self-supervised learning,'' \emph{IEEE TNNLS}, vol.~33, no.~6, pp. 2378--2392, 2021.

\bibitem{zhang2022reconstruction}
J.~Zhang, S.~Wang, and S.~Chen, ``Reconstruction enhanced multi-view contrastive learning for anomaly detection on attributed networks,'' \emph{arXiv}, 2022.

\bibitem{li2019specae}
Y.~Li, X.~Huang, J.~Li, M.~Du, and N.~Zou, ``Specae: Spectral autoencoder for anomaly detection in attributed networks,'' in \emph{CIKM}, 2019, pp. 2233--2236.

\bibitem{ding2019deep}
K.~Ding, J.~Li, R.~Bhanushali, and H.~Liu, ``Deep anomaly detection on attributed networks,'' in \emph{SDM}.\hskip 1em plus 0.5em minus 0.4em\relax SIAM, 2019, pp. 594--602.

\bibitem{roy2024gad}
A.~Roy, J.~Shu, J.~Li, C.~Yang, O.~Elshocht, J.~Smeets, and P.~Li, ``Gad-nr: Graph anomaly detection via neighborhood reconstruction,'' in \emph{WSDM}, 2024, pp. 576--585.

\bibitem{chen2023anomman}
L.-H. Chen, H.~Li, W.~Zhang, J.~Huang, X.~Ma, J.~Cui, N.~Li, and J.~Yoo, ``Anomman: Detect anomalies on multi-view attributed networks,'' \emph{Information Sciences}, vol. 628, pp. 1--21, 2023.

\bibitem{ren2024sslrec}
X.~Ren, L.~Xia, Y.~Yang, W.~Wei, T.~Wang, X.~Cai, and C.~Huang, ``Sslrec: A self-supervised learning framework for recommendation,'' in \emph{WSDM}, 2024, pp. 567--575.

\bibitem{fu2023multiplex}
C.~Fu, G.~Zheng, C.~Huang, Y.~Yu, and J.~Dong, ``Multiplex heterogeneous graph neural network with behavior pattern modeling,'' in \emph{SIGKDD}, 2023, pp. 482--494.

\bibitem{mcauley2013amateurs}
J.~J. McAuley and J.~Leskovec, ``From amateurs to connoisseurs: modeling the evolution of user expertise through online reviews,'' in \emph{WWW}, 2013, pp. 897--908.

\bibitem{rayana2015collective}
S.~Rayana and L.~Akoglu, ``Collective opinion spam detection: Bridging review networks and metadata,'' in \emph{SIGKDD}, 2015, pp. 985--994.

\bibitem{tang2023gadbench}
J.~Tang, F.~Hua, Z.~Gao, P.~Zhao, and J.~Li, ``Gadbench: Revisiting and benchmarking supervised graph anomaly detection,'' \emph{NeurIPS}, vol.~36, pp. 29\,628--29\,653, 2023.

\bibitem{ding2019interactive}
K.~Ding, J.~Li, and H.~Liu, ``Interactive anomaly detection on attributed networks,'' in \emph{WSDM}, 2019, pp. 357--365.

\bibitem{jin2021anemone}
M.~Jin, Y.~Liu, Y.~Zheng, L.~Chi, Y.-F. Li, and S.~Pan, ``Anemone: Graph anomaly detection with multi-scale contrastive learning,'' in \emph{CIKM}, 2021, pp. 3122--3126.

\bibitem{duan2023arise}
J.~Duan, B.~Xiao, S.~Wang, H.~Zhou, and X.~Liu, ``Arise: Graph anomaly detection on attributed networks via substructure awareness,'' \emph{IEEE TNNLS}, 2023.

\bibitem{zheng2021generative}
Y.~Zheng, M.~Jin, Y.~Liu, L.~Chi, K.~T. Phan, and Y.-P.~P. Chen, ``Generative and contrastive self-supervised learning for graph anomaly detection,'' \emph{IEEE TKDE}, vol.~35, no.~12, pp. 12\,220--12\,233, 2021.

\bibitem{pan2023prem}
J.~Pan, Y.~Liu, Y.~Zheng, and S.~Pan, ``Prem: A simple yet effective approach for node-level graph anomaly detection,'' in \emph{ICDM}.\hskip 1em plus 0.5em minus 0.4em\relax IEEE, 2023, pp. 1253--1258.

\bibitem{kipf2016variational}
T.~N. Kipf and M.~Welling, ``Variational graph auto-encoders,'' \emph{arXiv}, 2016.

\bibitem{fan2020anomalydae}
H.~Fan, F.~Zhang, and Z.~Li, ``Anomalydae: Dual autoencoder for anomaly detection on attributed networks,'' in \emph{ICASSP}.\hskip 1em plus 0.5em minus 0.4em\relax IEEE, 2020, pp. 5685--5689.

\bibitem{bandyopadhyay2020outlier}
S.~Bandyopadhyay, L.~N, S.~V. Vivek, and M.~N. Murty, ``Outlier resistant unsupervised deep architectures for attributed network embedding,'' in \emph{WSDM}, 2020, pp. 25--33.

\bibitem{chen2024boosting}
J.~Chen, G.~Zhu, C.~Yuan, and Y.~Huang, ``Boosting graph anomaly detection with adaptive message passing,'' in \emph{ICLR}, 2024.

\end{thebibliography}

\end{document}